\documentclass[aos,preprint]{imsart}

\RequirePackage[OT1]{fontenc}
\RequirePackage{amsthm,amsmath}
\RequirePackage[numbers]{natbib}
\RequirePackage[colorlinks,citecolor=blue,urlcolor=blue]{hyperref}

\usepackage[usenames, dvipsnames]{color}
\usepackage{amssymb, amsfonts}
\usepackage{bbm}

\usepackage{amsmath}
\usepackage{amsthm}

\usepackage[normalem]{ulem}
\usepackage{natbib}
\usepackage{url}

\usepackage{graphicx}
\usepackage{amssymb}
\usepackage{subfigure}

\usepackage{enumitem}

\newtheorem{theorem}{\vspace{.1in}Theorem}
\newtheorem{definition}[theorem]{Definition}
\newtheorem{lem}[theorem]{Lemma}

\newcommand{\MX}{\textsl{\textbf{X}}}

\newcommand{\cis}{\mbox{\tiny{CIS}}}

\arxiv{arXiv:0000.0000}

\startlocaldefs
\numberwithin{equation}{section}
\theoremstyle{plain}

\endlocaldefs

\begin{document}

\begin{frontmatter}
\title{Classification with Ultrahigh-Dimensional Features\thanksref{T1}}
\runtitle{Classification with Ultrahigh-Dimensional Features}
\thankstext{T1}{Footnote to the title with the ``thankstext'' command.}

\begin{aug}
\author{\fnms{Yanming} \snm{Li}\thanksref{m1}\ead[label=e1]{liyanmin@umich.edu}},
\author{\fnms{Hyokyoung} \snm{Hong}\thanksref{t2,m2}\ead[label=e2]{hhong@stt.msu.edu}},
\author{\fnms{Jian} \snm{Kang}\thanksref{t3,m1}
\ead[label=e3]{jiankang@umich.edu}},
\author{\fnms{Kevin} \snm{He}\thanksref{m1}
\ead[label=e4]{kevinhe@umich.edu}},
\author{\fnms{Ji} \snm{Zhu}\thanksref{m1}
\ead[label=e5]{jizhu@umich.edu}}
\and
\author{\fnms{Yi} \snm{Li}\thanksref{m1}
\ead[label=e6]{yili@med.umich.edu}}

\thankstext{t1}{Some comment}
\thankstext{t2}{Supported in part by the NSA Grant H98230-15-1-0260}
\thankstext{t3}{Supported in part by the NIH Grant 1R01MH105561}
\runauthor{Y. Li et al.}

\affiliation{University of Michigan\thanksmark{m1} and Michigan State University\thanksmark{m2}}

\address{Yanming Li\\
Department of Biostatistics,\\
University of Michigan \\
1415 Washington Heights \\
Ann Arbor, Michigan 48109 \\
E-mail:\ \printead*{e1}}

\address{Yi Li\\
Department of Biostatistics,\\
University of Michigan \\
1415 Washington Heights \\
Ann Arbor, Michigan 48109 \\
E-mail:\ \printead*{e6}\\}
\end{aug}

\begin{abstract}
Although much progress has been made in classification with high-dimensional features  \citep{Fan_Fan:2008, JGuo:2010, CaiSun:2014, PRXu:2014},
classification with ultrahigh-dimensional features,
wherein the features  much outnumber the sample size, defies most existing work.
This paper introduces  a  novel and computationally feasible multivariate screening and classification method for  ultrahigh-dimensional data. Leveraging inter-feature correlations, the proposed method enables detection of  marginally weak and sparse signals and recovery of the true informative feature set, and achieves asymptotic optimal misclassification rates.
We also show that the proposed procedure provides more powerful discovery boundaries compared to those in \citet{CaiSun:2014} and \citet{JJin:2009}.
The performance of the  proposed procedure  is evaluated using simulation studies and  demonstrated via classification of  patients with different post-transplantation renal functional types.
\end{abstract}



\end{frontmatter}

\section{Introduction}
{High-throughput data such as genomic, proteomic, microarray and neuroimaging data have now been routinely collected in many contemporary biomedical studies}, wherein the number of variables  far exceeds the number of observations. For example, our motivating kidney transplantation study \citep{FLENCHER:2004} {assayed} a total of 12,625 genes  from 62 tissue samples with 4 different post-transplantation kidney functional types.
{A key scientific interest lies in classifying} patients with different functional types based on their molecular  information, for the purpose of precision medicine.

Classification methods for high-dimensional data have been widely studied in recent years \citep{YGuo:2007, Fan_Fan:2008, DWitten:2009, JGuo:2010, ShaoJ:2011, DWitten:2011, JFan:2012, Qmai:2012, PRXu:2014}. However, most classification approaches  \citep{JGuo:2010, DWitten:2011, PRXu:2014}
 and software (penalizedLDA, lda in \texttt{R} \citep{CRANR}, etc.) use
penalized  methods,  which are not directly applicable to ultrahigh-dimensional cases due to  computational infeasibility.

\begin{figure}\label{fig:f1}
\centering
{
  \includegraphics[bb=1 16 440 233, clip, scale=0.8]{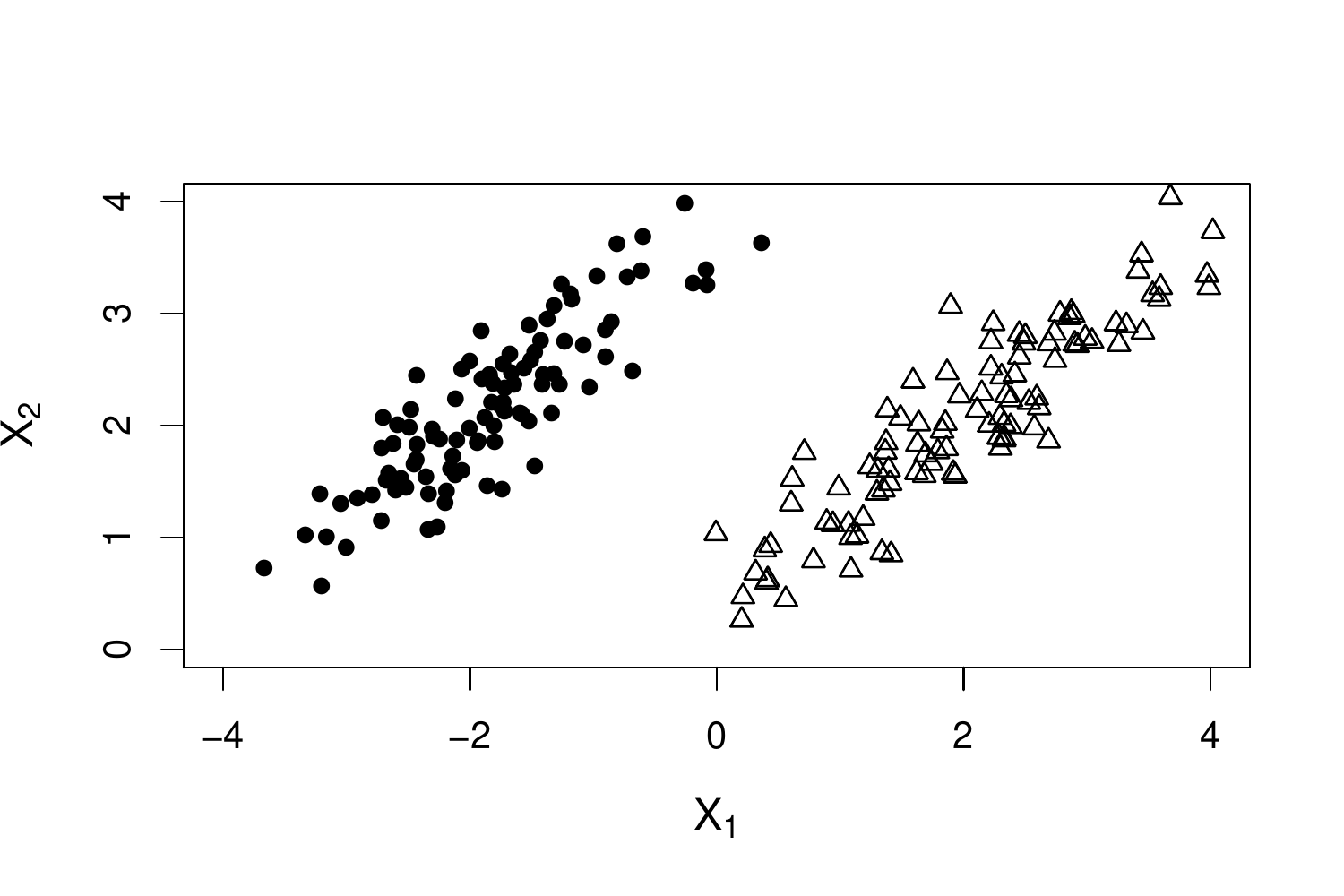}
  \caption{An illustrative example with two classes. Although $X_1$ is marginally informative, there is no single $X_1$ value that totally separates the two classes. However, one can easily find a line in the $X_1$-$X_2$ plane that completely separates the two classes. Therefore, even it is marginally non-informative, $X_2$ should be considered as a jointly informative feature.} \label{fig:2classes}
  }
\end{figure}

{
Furthermore, discriminant methods based on independence rules which  ignore correlations among features \citep{JLv:2008, RSong:2008}} have been widely practiced. This is partly because, as \citet{PBickel:2004} pointed out, the independence rule may perform better  in the high-dimensional case due to ill-conditioned high-dimensional covariance matrices and cumulative error arising from estimation. {However, this conclusion only holds for marginally strong signals.
Most current ultrahigh-dimensional screening methods \citep{Fan_Fan:2008, CaiSun:2014}  have assumed the independence rule when detect marginally strong signals.
In many real world applications, an independence rule is restrictive and frequently fails to detect marginally weak  features. \citet{PRXu:2014} showed that even though a feature is marginally non-discriminative, it could be informative when considered jointly with  other marginally differentiating features. For example,  Figure \ref{fig:f1} indicates that the best classifier  involves both features $X_1$ and $X_2$ while $X_2$   does not have any discriminant power on its own.}

{Table \ref{tab:HDmethods}, which summarizes popular modern high-dimensional classification methods. It reveals that weak signal detection has been less studied in the literature, even though detecting marginally weak signals has drawn recent attention\citep{Bodmer:2008, BLi:2008}.}
In this paper, we propose an ultrahigh-dimensional  screening and classification method, which accounts for inter-feature correlations and  enables detection of marginally weak or non-discriminant features that might be missed by marginal screening approaches. {We cast the proposed method in a unified algorithm for both ultrahigh-dimensional variable selection and classification motivated by the formulation of linear discriminant analysis (LDA). It contains a screening step that simultaneously selects informative features and estimates the parameters, as well as  a classification step.} We show that {the screening step  can  recover the informative features with a probability tending to $1$ and the classification step reaches an optimal misclassification rate even for weak signals.} Furthermore, it improves the discovery boundary defined in \citet{JJin:2009} and \citet{CaiSun:2014}. The procedure is computationally efficient and an \texttt{R} package, ``CIS-LDA," implementing the proposed method is also developed.

The rest of the paper is organized as follows. In Section 2, we introduce the covariance insured screening and classification method for ultrahigh-dimensional data. Section 3 develops theoretical properties of the proposed method. In Section 4, the performance of the proposed method is evaluated using simulation studies. We apply the proposed procedure to the renal transplantation data in Section 5 and Section 6 provides some discussions.

\begin{table}
\center \caption{Comparison between different high-dimensional classification methods}\label{tab:HDmethods}
\begin{tabular}{lcccc}
\hline
Methods & Handle & Consider & Able to & Identify pair\\
        & ultrahigh & inter-feature & detect &class-specific \\
        & dimension? & correlation? & marginally & features? \\
        & & & weak features?& \\
 \hline
   \citet{JGuo:2010} & \textbf{X} & \textbf{X} & \textbf{X}& $\boldsymbol\surd$ \\
   \citet{DWitten:2011}  & \textbf{X} & \textbf{X}& \textbf{X} & \textbf{X} \\
   \citet{Clemmensen:2011} & \textbf{X} & \textbf{X} & \textbf{X}&\textbf{X} \\
   \citet{JLv:2008} & $\boldsymbol\surd$ & \textbf{X}& \textbf{X}& \textbf{X} \\
   \citet{PRXu:2014} & \textbf{X} & $\boldsymbol\surd$ & $\boldsymbol\surd$ & $\boldsymbol\surd$ \\
   \citet{CaiSun:2014} & $\boldsymbol\surd$ & \textbf{X} & \textbf{X}&\textbf{X} \\
   \citet{TCaiWliu:2011} & \textbf{X}& $\boldsymbol\surd$& $\boldsymbol\surd$ & \textbf{X}\\
  \hline
\end{tabular}
\end{table}

\section{Covariance insured screening and classification}
\subsection{Notation}
Throughout the paper, we use bold upper-case letters to denote vectors and matrices. For a $p\times p$ matrix $\mathbf{A}$, denote by $\mathbf{A}'$ its transpose, by $\mathbf{A}_{\cdot j}$ its $j$th column vector, and by $A_{jj'}$ its entry on the $j$th row and $j'$th column, $1\leq j, j' \leq p$. Denote by $|S|$ the cardinality of a set $S$ and by $S^c$ the complement of $S$. 

Denote  by $\mathcal{G}(\mathcal{V}, \mathcal{E}; \mathbf{A})$ the graph induced by a $p\times p$ symmetric matrix $\mathbf{A}$, where the node set $\mathcal{V}=\{1, \cdots, p\}$ corresponds to the row indices of $\mathbf{A}$ and the edge set $\mathcal{E}$ contains all the  edges. An edge between nodes $j$ and $j'$  exists if and only if $A_{jj'}\neq 0$. For a subset $\mathcal{V}_l\subset \mathcal{V}$, denote by  $\mathbf{A}_l$ the principle submatrix of $\mathbf{A}$ confined on $\mathcal{V}_l$ and $\mathcal{E}_l$ the corresponding edge set. We say that the graph $\mathcal{G}(\mathcal{V}_l, \mathcal{E}_l, \mathbf{A}_l)$ is a connected component of $\mathcal{G}(\mathcal{V}, \mathcal{E}; {\mathbf{A}})$ if the following two conditions are satisfied: (i) any two nodes in $\mathcal{V}_l$ are connected by edges in $\mathcal{E}_l$ and (ii) for any node in $\mathcal{V}\cap\mathcal{V}_l^c$, there exists another node in $\mathcal{V}$ such that the two can not be connected by edges in $\mathcal{E}$.

For a symmetric matrix $\mathbf{A}$, denote by $tr(\mathbf{A})$ the trace of $\mathbf{A}$, denote by $\lambda_{\mbox{\scriptsize{min}}}(\mathbf{A})$ and $\lambda_{\mbox{\scriptsize{max}}}(\mathbf{A})$ the minimum and maximum eigenvalues of $\mathbf{A}$. Denote the operator norm and the Frobenius norm by $\|\mathbf{A}\|=\lambda_{\mbox{\scriptsize{max}}}^{1/2}(\mathbf{A}'\mathbf{A})$ and $\|\mathbf{A}\|_F=tr(\mathbf{A}'\mathbf{A})^{1/2}$, respectively.

\subsection{Method} We assume that there are $K$ distinct classes. Let $Y$ represent class label taking values in $\{1, \cdots, K\}$.
Denote by  $w_k=P(Y=k)>0$ for $k=1,\ldots, K$ such that $\sum_{k=1}^K w_{k}=1$.
 Let $\MX$ be a $p$-dimensional vector of features, of which the distribution  conditional on the corresponding class membership  follows a  multivariate Gaussian
\begin{equation}\label{eqn:eqn1}
 \MX|Y=k \sim N(\boldsymbol{\mu}_k, \boldsymbol{\Sigma}), \ \ \ \ k=1,\cdots, K,
\end{equation}
where $\boldsymbol{\mu}_k$ is a $p$-dimensional vector, $k=1,\cdots, K$ and $\boldsymbol{\Sigma}$ is a $p\times p$ positive definite matrix. Let $(Y_1, \MX_1), \cdots, (Y_n, \MX_n)$ be $n$ independent observations of $(Y, \MX)$. When $K=2$, (\ref{eqn:eqn1}) can be rewritten as
\begin{equation}\label{eqn:eqn1TwoCases}
\MX_i|Y_i \sim (2-Y_i) N(\boldsymbol{\mu}_1, \boldsymbol{\Sigma}) + (Y_i-1) N(\boldsymbol{\mu}_2, \boldsymbol{\Sigma}), \ \ \ \ i=1,\cdots, n .
\end{equation}

{For the ease of presentation,  we assume $K=2$ throughout}.  We note that the  theoretical properties, implementation, and our \texttt{R} package can all  be readily extended to the $K>2$ case.

{We recast (\ref{eqn:eqn1TwoCases}) in the framework of rare/weak feature models  \citep{JJin:2009}, where $\boldsymbol{\mu_2}$ is only different from $\boldsymbol{\mu_1}$ in $\epsilon$ fraction of features. } Specifically, let $I_1, I_2, \cdots, I_p$ be $p$ samples from a Bernoulli($\epsilon$) distribution, and
\begin{equation}\label{eqn:MDif}
\mu_{2j}=\mu_{1j}+ \delta_j I_j, \ \ 1\leq j\leq p,
\end{equation}
where $\mu_{kj}$ denotes the $j$th coordinate of $\boldsymbol{\mu}_k$, $k=1, 2$. $\boldsymbol{\delta}=(\delta_1, \cdots, \delta_p)'$ and $\delta_j\neq 0$ for all $j$. Equation (\ref{eqn:MDif}) assigns the mean difference between the two classes on features with $I_j=1$. Set $\tau=\min\{|\delta_j|, I_j=1, 1\leq j \leq p\}$. {Therefore, $\tau$  controls the strength of signals and $\epsilon$ controls the rareness  of marginally differentiable features.}
\citet{JJin:2009} and \citet{CaiSun:2014} showed that in the ultrahigh-dimensional case with both small $\tau$ and $\epsilon$, it is impossible to  discriminate one class from the other. Following \citet{JJin:2009}, let $\tau=\tau_p=\sqrt{r\log p}$ for some $0 <r<1$ and $\epsilon=\epsilon_p=p^{-\beta}$ for some $0<\beta<1$.
Then $\tau$ and $\epsilon$ are controlled by $r$ and $\beta$, respectively.
{As $r$ gets closer to $0$  the signal becomes weaker, while as $\beta$ gets closer to $1$  the signal becomes sparser.}

Next we outline a  new  covariance-insured screening (CIS) and classification method that utilizes the {feature} correlation structures  to detect the signals that are marginally uninformative but jointly informative ({MUJI}). Denote  by  $\boldsymbol{\Omega}=\boldsymbol{\Sigma}^{-1}$  the precision matrix, where  $\Omega_{jj'}=0$ if and only if that the $j$th and $j'$th ($1\leq j\neq j' \leq p$) variables are conditionally uncorrelated given the rest of the variables. It was shown in \citep{PRXu:2014} that a sufficient and necessary condition
 for variable $j$ to be non-informative in differentiating the two classes is
 \begin{equation}\label{eqn:iff}
 \sum_{j'=1}^p \Omega_{jj'}(\mu_{1j'}-\mu_{2j'})=0. \tag{Condition A}
 \end{equation}

\begin{figure}\label{fig:conn_Omega}
\centering
{ \includegraphics[scale=0.5]{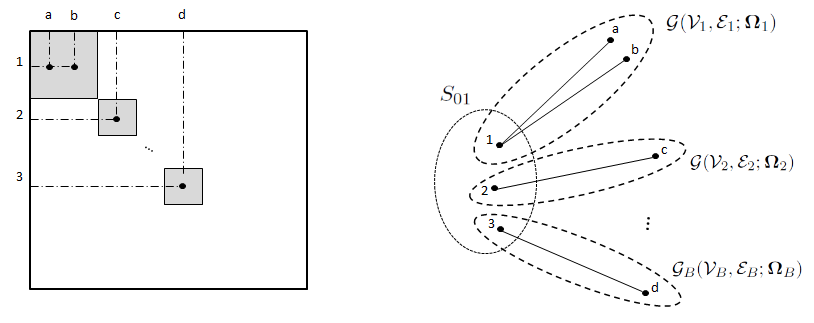}
  \caption{An illustrative example of marginally informative signals and their connected components in $\mathcal{G}(\mathcal{V}, \mathcal{E}; \boldsymbol{\Omega})$. Denote by $S_{01}$ the set of marginally informative signals. Features 1, 2,  and 3 are marginally informative signals and features a, b, c, and d are MUJI signals. The left panel shows the diagonal block structure of $\boldsymbol{\Omega}$. Each block contains at least one marginally informative feature and other non-informative features in their connected component. The right panel illustrates the corresponding graph structure in
  $\mathcal{G}(\mathcal{V}, \mathcal{E}; \boldsymbol{\Omega})$.} \label{fig:connOmegaBlock}
  }
\end{figure}

 Accurately estimating the precision matrix $\boldsymbol{\Omega}$  involves inverting a large square matrix and requires $O(p^3)$ computational steps. However, under the sparsity assumption of $\boldsymbol{\Omega}$, it is possible to estimate $\sum_{j'=1}^p \Omega_{jj'}(\mu_{1j'}-\mu_{2j'})$ without fully uncovering the whole $\boldsymbol{\Omega}$ matrix. {Specifically, assume that in each row of $\boldsymbol{\Omega}$, the fraction of its nonzero entries is at most $\vartheta=\vartheta_p=p^{-\gamma}$ for $0<\gamma <1$.
 Hence,  }each marginally informative signal is partially correlated with at most  $\vartheta p$ number of the MUJI signals.
 Under the sparsity assumption, $\mathcal{G}(\mathcal{V}, \mathcal{E}; {\boldsymbol{\Omega}})$ is a $(\vartheta p)$-sparse graph in the sense that the degree of each node $\leq \vartheta p$. Particularly, denote by $S_{01}=\{j : \mu_{1j}-\mu_{2j}\neq 0, j=1,\cdots,p \}$, then each feature in $S_{01}$ is partially correlated with at most $\vartheta p$ features. Suppose the graph $\mathcal{G}(\mathcal{V}, \mathcal{E}; {\boldsymbol{\Omega}})$ admits a decomposition of $B$ connected components and denote by $\boldsymbol{\Omega}_1, \cdots, \boldsymbol{\Omega}_B$ the {precision matrices} in $\boldsymbol{\Omega}$ corresponding to those connected components, i.e.
\[\mathcal{G}(\mathcal{V}, \mathcal{E}; {\boldsymbol{\Omega}})=\bigcup_{l=1}^B \mathcal{G}(\mathcal{V}_l, \mathcal{E}_l; \boldsymbol{\Omega}_l),\] where $\mathcal{G}(\mathcal{V}_l, \mathcal{E}_l; \boldsymbol{\Omega}_l)$ is the sub-graph corresponds to the $l$th connected component. \ref{eqn:iff} reveals that if a marginally non-informative feature, say $j$, is jointly informative, there must exist a feature $j'\neq j$ such that $\mu_{1j'}-\mu_{2j'}\neq 0$ and $\Omega_{jj'}\neq 0$. That is, feature $j$ has to be connected to some marginally informative feature. This implies that, in order to detect the MUJI signals, one only needs to focus on the connected components that contains at least one marginally informative features in $S_{01}$. See Figure \ref{fig:conn_Omega} for a graphical illustration.


Under the sparsity assumption, the connected components and the graphic structure of $\boldsymbol{\Omega}$ can be inferred by thresholding the sample covariance matrix $\hat{\boldsymbol{\Sigma}}$. Denote by $\tilde{\boldsymbol{\Sigma}}^{\alpha}$ the thresholded sample covariance matrix with a threshold $\alpha$, where $\tilde{\Sigma}^{\alpha}_{jj'}=\hat{\Sigma}_{jj'}\mathbbm{1}\{|\hat{\Sigma}_{jj'}|\geq \alpha\}$, $1\leq j, j'\leq p$ with $\mathbbm{1}(\cdot)$ being the indicator function. Denote by $\mathcal{G}(\mathcal{V}, \tilde{\mathcal{E}}; \tilde{\boldsymbol{\Sigma}}^{\alpha})$ the graph corresponding to $\tilde{\boldsymbol{\Sigma}}^{\alpha}$ and suppose it can be decomposed into $\tilde{B}$ connected components
\[\mathcal{G}(\mathcal{V}, \tilde{\mathcal{E}}; \tilde{\boldsymbol{\Sigma}}^{\alpha})=\bigcup_{l=1}^{\tilde{B}} \mathcal{G}(\tilde{\mathcal{V}}_l, \tilde{\mathcal{E}}_l; \tilde{\boldsymbol{\Sigma}}^{\alpha}_l)\] with $\tilde{\mathcal{V}}_l$, $\tilde{\mathcal{E}}_l$, $\tilde{\boldsymbol{\Sigma}}^{\alpha}_l$ being the node set, edge set and the sub-matrix in $\tilde{\boldsymbol{\Sigma}}^{\alpha}$ corresponding to the $l$th connected component, $l=1, \cdots, \tilde{B}$. For two symmetric matrices $\mathbf{A}$ and $\tilde{\mathbf{A}}$, we say that graph $\mathcal{G}(\mathcal{V}, \mathcal{E}; \mathbf{A})$ belongs to graph $\mathcal{G}(\tilde{\mathcal{V}}, \tilde{\mathcal{E}}; \tilde{\mathbf{A}})$, denote by $\mathcal{G}(\mathcal{V}, \mathcal{E}; \mathbf{A}) \subseteq \mathcal{G}(\tilde{\mathcal{V}}, \tilde{\mathcal{E}}; \tilde{\mathbf{A}})$, if $\mathcal{V}\subseteq \tilde{\mathcal{V}}$ and $\mathcal{E}\subseteq \tilde{\mathcal{E}}$.

\begin{lem}\label{lem:subgraph}
Let $\mathcal{G}(\mathcal{V}_j, \mathcal{E}_j; \boldsymbol{\Omega}_j)$ be the connected component of a marginally informative feature $j$ in $\mathcal{G}(\mathcal{V}, \mathcal{E}; \boldsymbol{\Omega})$, and $\mathcal{G}(\tilde{\mathcal{V}}_j, \tilde{\mathcal{E}}_j; \tilde{\boldsymbol{\Sigma}}^{\alpha}_j)$ be the connected component of feature $j$ in $\mathcal{G}(\mathcal{V}, \tilde{\mathcal{E}}; \tilde{\boldsymbol{\Sigma}}^{\alpha})$, then for sufficiently large $n$ and $\alpha=O(n^{-\kappa})$, we have
\[P\left(\mathcal{G}(\mathcal{V}_j, \mathcal{E}_j; \boldsymbol{\Omega}_j) \subseteq \mathcal{G}(\tilde{\mathcal{V}}_j, \tilde{\mathcal{E}}_j; \tilde{\boldsymbol{\Sigma}}^{\alpha}_j)\right) \geq  1-C_1\exp(-C_2 n^{1-2\kappa}),\]
where $C_1$, $C_2$ are some positive constants and $0<\kappa<1/2$.
\end{lem}
The proof of Lemma \ref{lem:subgraph} is similar to Theorem 1 in \citep{SLuo:2014} and is provided in the supplemental material.

\textsc{Remark.} {The efficiency gain on estimating $\boldsymbol{\Omega}$ stems} from that: (i) instead of solving an ultrahigh-dimensional graphic lasso problem, it computes the connected components of the thresholded sample covariance matrix, reducing computational cost;  (ii) it only needs  detection of  the connected components of the marginally informative ones. Lemma \ref{lem:subgraph} guarantees the efficacy of the proposed method on reconstructing the connected component structure.

\subsection{covariance-insured screening and classification steps}

Given the data set $\{\mathbb{Y}, \mathbf{\mathbb{X}}\} = \{Y_i, \mathbf{X}_i\}_{i=1}^n$, denote by $n_k$ the number of samples in class $k$, $k=1, 2$.
{Denote by $X_{ij}$ the value of feature $j$ for sample $i$ and $\bar{X}_{\cdot j}^{(k)}$ the mean of feature $j$ in class $k$, $k=1, 2$. We further detail the proposed procedures that consist of screening and classification steps.
\medskip
\begin{itemize}
\item[]\emph{\underline{Screening step}}\\

 \item[1.] Standardize $\mathbb{X}$ columnwise.
 \item[2.] Select the set $S^{0,\tau}$ consisting of the features $\mathbf{X}_{\cdot j}$ satisfying
 \begin{equation}
 \left|\bar{X}_{\cdot j}^{(1)} - \bar{X}_{\cdot j}^{(2)}\right|= \left|\frac{1}{n_1}\sum_{Y_i=1}X_{ij}- \frac{1}{n_2}\sum_{Y_i=2} X_{ij}\right| > \tau,
 \end{equation}
 where $\tau$ is a thresholding parameter that controls the strength of the marginal effects.
 \item[3.] For a given threshold $\alpha$, construct a sparse estimate of the covariance matrix $\tilde{\boldsymbol{\Sigma}}^{\alpha}$ by setting $\tilde{\Sigma}^{\alpha}_{jj'}=\hat{\Sigma}_{jj'} \mathbbm{1}\{|\hat{\Sigma}_{jj'}| \geq \alpha\}$, $1\leq j, j' \leq p$.
  \item[4.] For each feature $j$ in $S^{0,\tau}$, detect its connected component in $\tilde{\boldsymbol{\Sigma}}^{\alpha}$. Denote by $S_j$ the corresponding index set. Suppose the first $k(\alpha, \tau)$ components are the smallest number of components such that $S^{0,\tau} \subset \cup_{g=1}^{k(\alpha, \tau)} S_g$. {Compute the precision matrix of the first $k(\alpha, \tau)$ components, $\hat{\boldsymbol{\Omega}}^{\alpha, \tau}= \mbox{diag}((\tilde{\boldsymbol{\Sigma}}^{\alpha}_{S_1})^{-1}, \cdots, (\tilde{\boldsymbol{\Sigma}}^{\alpha}_{S_{k(\alpha, \tau)}})^{-1})$.}
  \item[5.] Evaluate \ref{eqn:iff} and select informative features by
  \begin{equation*}
  S^{\cis}(\alpha, \tau; \nu_n)=\left\{1 \leq j \leq p: \left|\sum_{j'\in \cup_{g=1}^{k(\alpha, \tau)} S_g} \hat{\Omega}^{\alpha, \tau}_{jj'}(\bar X_{\cdot j'}^{(1)}-\bar X_{\cdot j'}^{(2)})\right| \geq \nu_n \right\}
  \end{equation*}
  for some pre-specified $\nu_n$. In our numerical studies, we rank the magnitude of the estimated \ref{eqn:iff}: $IS_j=|\sum_{j'\in \cup_{g=1}^{k(\alpha, \tau)} S_g} \hat{\Omega}^{\alpha, \tau}_{jj'}(\bar X_{\cdot j'}^{(1)}-\bar X_{\cdot j'}^{(2)})|$ and select up to the first $n$ features, where $n$ is the sample size. $IS_j$ refers to the importance score of predictor $j$.
\medskip

\textsc{Remark.}  The optimal tuning parameter values $\tau$ and $\alpha$ are selected through cross-validation. Resampling strategies such as stability selection \citep{Meinshausen:2010} can be applied to reduce the variation from random sampling.
\medskip

\item[6.]  \emph{\underline{Post-screening classification}}\\

{A new observation $\mathbf{X}_{\mbox{\scriptsize{new}}}$ is to be classified  with} the decision rule $\mathbbm{1}(\hat\delta^{\cis}(\alpha, \tau, \nu_n)\geq 0)$ with $\hat\delta^{\cis}(\alpha, \tau; \nu_n)=(\mathbf{X}_{\mbox{\scriptsize{new}}}-\hat{\boldsymbol{\mu}}^{\cis})'\hat{\boldsymbol{\Omega}}^{\cis}(\hat{\boldsymbol{\mu}}^{\cis}_1-\hat{\boldsymbol{\mu}}^{\cis}_2)$. Here $\hat{\boldsymbol{\mu}}_1=\sum_{\{i: Y_i=1\}} \mathbf{X}_i/n_1$, $\hat{\boldsymbol{\mu}}_2=\sum_{\{i: Y_i=2\}} \mathbf{X}_i/n_2$, $\hat{\boldsymbol{\mu}}=(\hat{\boldsymbol{\mu}}_1+ \hat{\boldsymbol{\mu}}_2)/2$, $\hat{\boldsymbol{\Omega}}=\hat{\boldsymbol{\Omega}}^{\alpha, \tau}$ and superscript ``CIS" restricts features to $S^{\mbox{\tiny{CIS}}}(\alpha, \tau; \nu_n)$.
  \end{itemize}
\medskip

{\textsc{Remark.} Notice that 2-4 in the screening step are mainly thresholding the marginal mean difference and the correlation parameters. They estimate the parameters needed in constructing the screening statistics. And 5 in the screening step does the job of variable selection based on \ref{eqn:iff}. Even though 2-4 can identify marginally informative signals and their connected components, but they do not select MUJI signals by themselves without 5.}

{At step 4 of screening, for each feature $j\in S^{0,\tau}$, we look for  $j'$ such that $\tilde{\Sigma}^{\alpha}_{jj'}$ is relatively large. For each of those $j'$s, we then recursively look for its connected features till all the features in the connected component of feature $j$ are found.} Other connected component labeling algorithms such as those listed in \citep{Shapiro:2002} can be employed. For a specific $\alpha$ and some $g\in \{1, \cdots, k(\alpha, \tau)\}$, it is possible that $\tilde{\boldsymbol{\Sigma}}^{\alpha}_g$ is of a greater dimension than the sample size and calculating $(\tilde{\boldsymbol{\Sigma}}^{\alpha}_g)^{-1}$ is by itself a complex high-dimensional graphical model problem. {In such cases, we consider an $m$-depth connected subgraph, which contains features connecting to the feature $j$ through at most $m$ edges in the graph corresponding to $\tilde{\boldsymbol{\Sigma}}^{\alpha}_g$. Denote the node set of such a connected subgraph by $S_g^{(m)}$.} We then set $\tilde{\boldsymbol{\Sigma}}^{\alpha}_g$ and $(\tilde{\boldsymbol{\Sigma}}^{\alpha}_g)^{-1}$ to be the covariance and precision matrices of features in $S_g^{(m)}$ only. {This is similar to the $m$-sized subgraph in \citet{Tracy:2014}.} With a proper choice of $m$, one can avoid inverting a high-dimensional covariance matrix while involving essentially all the highly correlated features. The  impact of the different sizes of m on screening and classification is explored in Section 4. {Using different depth $m$ for different connected component is}, to some extent, equivalent to adaptively thresholding the corresponding correlation blocks for different features in $S^{0,\tau}$.

\section{Theoretical properties}

Let $S_0=\{j: \sum_{j'=1}^p \Omega_{jj'}(\mu_{1j'}-\mu_{2j'})\neq 0, j=1,\cdots,p \}$ be the true informative set. Then $S_0=S_{01} \cup S_{02}$, where $S_{01}=\{j : \mu_{1j}-\mu_{2j}\neq 0, j=1,\cdots,p \}$ and $S_{02}=\{j: \mu_{1j}-\mu_{2j}= 0, \sum_{j'=1}^p \Omega_{jj'}(\mu_{1j'}-\mu_{2j'})\neq 0, j=1,\cdots,p \}$. And $S^c_0=\{j: \sum_{j'=1}^p \Omega_{jj'}(\mu_{1j'}-\mu_{2j'})=0, j=1,\cdots,p \}$.

Let $S^{\cis}(\alpha, \tau; \nu_n)$ be the index set of features selected by the {CIS screening step} with thresholding parameters $(\alpha, \tau; \nu_n)$. Let $C_{h,p}=\max_{1\leq i\leq p}\sum_{1\leq j\leq p} |\Sigma_{ij}|^h$, where $ 0<h<1$ is a constant not depending on $p$. Note that $C_{h,p}$ can be used as a measure of the overall sparsity of $\Sigma$ \citep{Bickel_Levina:2008, ShaoJ:2011, Fan_Liao:2011}. Then under the following assumptions, $S^{\cis}(\alpha, \tau; \nu_n)$ can uncover the true informative set with probability tending to $1$.

\noindent(A1) $C_{h,p} <M <\infty$ for some positive $M$; \\
\noindent(A2) $\log p =O(n^\xi)$ for some $\xi\in (0,1)$; \\
\noindent(A3) $c_{p}=\min_{1\leq j,\,j'\leq p} |\Sigma_{ij}|=O(n^{-\kappa})$ for some $0<\kappa <1/2$ {such that $2\kappa+\xi<1$ for $\xi$ in (A2)}; \\
\noindent(A4) there exist positive constants $\kappa_1$ and $\kappa_2$ such that $0< \kappa_1 <\lambda_{\mbox{\scriptsize{min}}}(\boldsymbol{\Omega}) \leq  \lambda_{\mbox{\scriptsize{max}}}(\boldsymbol{\Omega}) < \kappa_2 < \infty$; \\
\noindent(A5) $\tau=O(\sqrt{r\log p})$, $\alpha=O(\sqrt{\log p/n})$ and $\nu_n=O\left(n\sqrt{r}/ \sqrt{\log p}\right)$;\\
\noindent(A6) $\beta+\gamma>1$, i.e. both the marginally informative signals and the MUJI signals are sparse enough.

\begin{theorem}\label{prop:FNC}
\textbf{(Sure Screening Property)} Under assumptions (A1)-(A5), for any $\epsilon>0$,
\begin{equation}\label{eqn:FNC}
P(|S_0 \cap S^{\cis}(\alpha, \tau; \nu_n)|\geq (1-\epsilon)|S_0|) \rightarrow 1 \mbox{ as } n\rightarrow \infty.
\end{equation}
\end{theorem}
\begin{theorem}\label{prop:FPC}
\textbf{(False Positive Control Property)} Under assumptions (A1)-(A6), for any $\zeta_n = O(n^r)$, we have
\begin{equation}\label{eqn:FPC}
P(|S^{\cis}(\alpha, \tau; \nu_n)\cap S^c_0| \leq \zeta_n^{-1} |S^c_0| ) \rightarrow 1 \mbox{ as } n\rightarrow \infty.
\end{equation}
\end{theorem}

\textsc{Remark.} Theorems \ref{prop:FNC} and \ref{prop:FPC} ensure that with probability tending to $1$, screening step of the CIS method captures all and only the informative features.

Let $R_{\mbox{\scriptsize{OPT}}}$ denote the misclassifiction rate of the oracle rule or the optimal rule, which assumes that the parameters $\boldsymbol{\mu}_1$, $\boldsymbol{\mu}_2$ and $\boldsymbol{\Sigma}$ are known \citep{McLachlan:2004}. Under model (\ref{eqn:eqn1}), it can be shown that $R_{\mbox{\scriptsize{OPT}}}=\Phi(-\Delta_p/2)$, where $\Delta_p=\sqrt{\boldsymbol{\delta}'\boldsymbol{\Sigma}^{-1}\boldsymbol{\delta}}$ \citep{McLachlan:2004} with $\boldsymbol{\delta}=\boldsymbol{\mu}_1-\boldsymbol{\mu}_2$ and $\Phi$ is the standard normal cumulative distribution function.
Consider the following definitions introduced in \citet{ShaoJ:2011}.

\begin{definition}
Let $\psi$ be a classification rule with the conditional misclassification rate $R_\psi(\mathbf{X})$, given the training sample $\mathbf{X}$.
\begin{itemize}[leftmargin=*]
\item[] (D3.1) $\psi$ is asymptotically optimal if $R_\psi(\mathbf{X})/R_{\mbox{\scriptsize{OPT}}}\rightarrow_P 1$;
\item[] (D3.2) $\psi$ is asymptotically sub-optimal if $R_\psi(\mathbf{X})-R_{\mbox{\scriptsize{OPT}}}\rightarrow_P 0$.
\end{itemize}
\end{definition}

For any thresholding parameters $\tau$ and $\alpha$ of the order specified in assumption (A5), denote by $\hat{\boldsymbol{\delta}}=\hat{\boldsymbol{\mu}}_1-\hat{\boldsymbol{\mu}}_2$ and by $\hat{\boldsymbol{\Omega}}^{\alpha, \tau}$  the estimated $p\times p$ precision matrix from the CIS screening step (fill the entries with zeros if $|\cup_{g=1}^{k(\alpha, \tau)} S_g| < p$). We first consider a classification rule using all the features: $\mathbbm{1}(\hat\psi^{\alpha,\tau}(\mathbf{X})\geq 0)$ with $\hat\psi^{\alpha,\tau}(\mathbf{X})=(\mathbf{X}-\hat{\boldsymbol{\mu}})'\hat{\boldsymbol{\Omega}}^{\alpha, \tau}\hat{\boldsymbol{\delta}}$ and $\hat{\boldsymbol{\mu}}=(\hat{\boldsymbol{\mu}}_1+\hat{\boldsymbol{\mu}}_2)/2$. Its misclassification rate is given by \citep{Fan_Fan:2008}:
\begin{equation}\label{eqn:MCerror}
R_{\cis} \equiv \frac{1}{2} \sum_{k=1}^2 \Phi\left(-\frac{ (\boldsymbol{\mu}_k- \hat{\boldsymbol{\mu}})'\hat{\boldsymbol{\Omega}}^{\alpha, \tau} \hat{\boldsymbol{\delta}}}{\sqrt{\hat{\boldsymbol{\delta}}' \hat{\boldsymbol{\Omega}}^{\alpha, \tau} \boldsymbol{\Sigma} \hat{\boldsymbol{\Omega}}^{\alpha, \tau} \hat{\boldsymbol{\delta}}}}\right).
\end{equation}
We further assume the following:

\noindent (A7) There exist positive constants $c_1$ and $c_2$ such that $0 < c_1 \leq \min\{n_1/n, n_2/n\} \leq \max\{n_1/n, n_2/n\} \leq c_2$; \\
\noindent (A8) Assume that $a_n\equiv p^{\frac{3-\beta-\gamma}{2}}/(n\Delta_p^2)\rightarrow 0$;\\
\noindent (A9) Let $b_n$ be the number of features with nonzero mean differences between two classes. Then $b_n^{1/2}/(\Delta_pn^{1/2})\stackrel{P}{\rightarrow}0$; \\
\noindent (A10) Let $\rho_n =M_0 C_{h,p}( n^{-1}\log p )^{(1-h)/2}$ for some constant $M_0$ and $0<h<1$. Then $\rho_n^{1/2}\rightarrow 0$.

\begin{theorem}\label{prop:Aymp_OPT} (\textbf{Asymptotic Misclassification Rate}) Under the assumptions (A7)-(A10), {when ignoring the effect of variable selection based on \ref{eqn:iff} in the screening step and constructing a classification rule just using the parameter estimators $\hat{\boldsymbol{\delta}}$ and $\hat{\boldsymbol{\Omega}}^{\alpha, \tau}$, we have:}
\begin{itemize}[leftmargin=*]
  \item[] (T\ref{prop:Aymp_OPT}.1) $R_{\cis} = \Phi(-[\sqrt{p^{1-\beta-\gamma}}+O_p(a_n)]\Delta_p/2)$.
  \item[] (T\ref{prop:Aymp_OPT}.2) If $\sqrt{p^{1-\beta-\gamma}}\Delta_p$ is bounded, the $R_{\cis}$ is asymptotically optimal.
  \item[] (T\ref{prop:Aymp_OPT}.3) If $\sqrt{p^{1-\beta-\gamma}}\Delta_p \rightarrow \infty$, the $R_{\cis}\rightarrow 0$ and is asymptotically sub-optimal.
\end{itemize}
\end{theorem}

{Theorem \ref{prop:Aymp_OPT} reveals that in ultrahigh-dimensional settings, when not combining with a variable selection method, building a decision rule simply based on all features can lead to a misclassification error no better than a mere random guess.}
\citet{Fan_Fan:2008} showed that in the case of uncorrelated features, for any classifier $\hat\psi$, the misclassification rate is $R(\hat\psi) \sim 1- \Phi\left(C \sqrt{n/p} \Delta_p\right)$ for some constant $C$. When the signal strength $\Delta_p$ is not strong enough to balance out the increasing dimensionality $p$, $\sqrt{n/p} \Delta_p \rightarrow 0$ as $n\rightarrow \infty$ and thus $R(\hat\psi)\stackrel{\scriptsize{P}}{\rightarrow} \frac{1}{2}$. Similarly for the proposed CIS classification, when all the features are used, according to Theorem \ref{prop:Aymp_OPT}, the signal strength $\Delta_p$ needs to be strong enough to balance out $\sqrt{p^{\beta+\gamma-1}}$ to achieve asymptotic optimal and zero misclassification rates. {When $\beta+\gamma=1$, Theorem \ref{prop:Aymp_OPT} reduces to Theorem 3 in \citep{ShaoJ:2011}, where sparsity is only assumed for marginally informative features, but not for MUJI features.}

However, according to Theorems \ref{prop:FNC} and \ref{prop:FPC}, the proposed CIS screening step retains only truly informative predictors $S_0$ with large probability. If we base the post-screening classification only on $S^{\cis}(\alpha, \tau; \nu_n)$, it can achieve better classification performance, because $\Delta_p$ then only needs to be strong enough to balance out $\sqrt{|S^{\cis}(\alpha, \tau; \nu_n)|}\ll p$ to yield an asymptotic zero and optimal post-screening misclassification rate.

Define the CIS post-screening classifier as $\mathbbm{1}(\hat\psi^{\mbox{\scriptsize{CIS\_PSC}}}\geq 0)$, where $ \hat\psi^{\mbox{\scriptsize{CIS\_PSC}}}=(\mathbf{X}^{\cis}-\hat{\boldsymbol{\mu}}^{\cis})'\hat{\boldsymbol{\Omega}}^{\cis}\hat{\boldsymbol{\delta}}^{\cis}$ with superscript CIS restricts the features to the set $S^{\cis}(\alpha, \tau; \nu_n)$ obtained from the CIS screening step. Then its misclassifcation rate is
\[R_{\mbox{\scriptsize{CIS\_PSC}}} \equiv \frac{1}{2} \sum_{k=1}^2 \Phi\left(-\frac{ (\boldsymbol{\mu}^{\cis}_k- \hat{\boldsymbol{\mu}}^{\cis})'\hat{\boldsymbol{\Omega}}^{\cis} \hat{\boldsymbol{\delta}}^{\cis}}{\sqrt{\hat{\boldsymbol{\delta}}^{\cis '} \hat{\boldsymbol{\Omega}}^{\cis} \hat{\boldsymbol{\Omega}}^{\cis} \boldsymbol{\Sigma}^{\cis} \hat{\boldsymbol{\Omega}}^{\cis}  \hat{\boldsymbol{\delta}}^{\cis}}}\right).
\]

\begin{theorem}\label{prop:postMR}
\textbf{(Post-screening Misclassification Rate)} Under the same conditions in Theorem \ref{prop:Aymp_OPT}, when classifying $\mathbf{X}_{\mbox{\scriptsize{new}}}$ based on features selected from the screening step, we have
\begin{itemize}[leftmargin=*]
\item[] (T\ref{prop:postMR}.1) $R_{\mbox{\scriptsize{CIS\_PSC}}} = \Phi(-[1+O_p(a_n)]\Delta_p/2)$.
\item[] (T\ref{prop:postMR}.2) If $\Delta_p$ is bounded, then $R_{\mbox{\scriptsize{CIS\_PSC}}}$ is asymptotically optimal.
\item[] (T\ref{prop:postMR}.3) If $\Delta_p \rightarrow \infty$, then $R_{\mbox{\scriptsize{CIS\_PSC}}}\rightarrow 0$ and is asymptotically sub-optimal.
\end{itemize}
\end{theorem}

Now let $S_\psi$ be the informative feature set discovered by a decision rule $\psi$. In order to characterize the phase transition in terms of optimal discovery, \citet{CaiSun:2014} defined the discovery boundary that satisfies the following conditions:
\begin{itemize}
\item[(C1)] The false positive rate is vanishingly small, i.e. $E(|S_\psi\cap S_0^c|/E(|S_\psi|)) \rightarrow 0$;
\item[(C2)] A non-empty discovery set is constructed with high probability, i.e. $P(|S_\psi|\geq 1)\rightarrow 1$.
\end{itemize}
Assuming that the marginally informative features have a common marginal variance $\sigma^2$ and marginally uninformative ones have a marginal variance $1$, \citet{CaiSun:2014} derived the discovery boundary $d(\beta)$ that divides the $\beta$-$r$ plane into discoverable and non-discoverable areas:
\begin{itemize}
\item For $\sigma=1$, $d(\beta) = (1-\sqrt{1-\beta})^2$;
\item For $0 <\sigma<1$, $d(\beta) = \left\{
                          \begin{array}{ll}
                            (1- \sigma\sqrt{1-\beta})^2, & \hbox{ if } 1-\sigma^2 <\beta<1 \\
                            (1-\sigma^2)\beta, & \hbox{ if } 0 < \beta\leq 1-\sigma^2;
                          \end{array}
                        \right.$
\item For $\sigma>1$, $d(\beta) = \left\{
                              \begin{array}{ll}
                               (1-\sigma\sqrt{1-\beta})^2, & \hbox{ if } 1-1/\sigma^2 <\beta<1 \\
                                0, & \hbox{ if }  0 < \beta <1-1/\sigma^2.
                              \end{array}
                            \right.$
\end{itemize}

Specifically, they showed that (a) If $r>d(\beta)$, it is possible to find a classification rule $\psi$ that fulfills (C1) and (C2) simultaneously. (b) If $r<d(\beta)$, it is impossible to find a classification rule $\psi$ that fulfills (C1) and (C2) simultaneously. A similar detection boundary about reliably detecting any existing signals was defined in \citep{JJin:2009}:
\[d_{\mbox{\scriptsize{det}}}(\beta)=\left\{
                                          \begin{array}{ll}
                                            0, & 0<\beta\leq 1/2; \\
                                            \beta-1/2, &  1/2<\beta \leq 3/4;\\
                                            (1-\sqrt{1-\beta})^2, & 3/4 <\beta<1.
                                          \end{array}
                                        \right.
\]

The discovery boundaries $d(\beta)$ coincide with $d_{\mbox{\scriptsize{det}}}(\beta)$ on $3/4 <\beta<1$. And $d(\beta) > d_{\mbox{\scriptsize{det}}}(\beta)$ on $0 <\beta \leq 3/4$. For $d_{\mbox{\scriptsize{det}}}(\beta)<r<d(\beta)$, one can detect the existence of signals reliably but it is impossible to separate any signal from noise \citep{CaiSun:2014}.

 However, when accounting for inter-feature correlations, we show in proposition \ref{prop:prop6} that the discovery boundary can be improved, and  the marginally uninformative features that {would have been} undiscoverable in \citep{CaiSun:2014} and \citep{JJin:2009} can become ``discoverable."
{Though finding the exact optimal discovery bound for model \eqref{eqn:eqn1TwoCases} is beyond the scope of this paper,  we do provide an upper bound of the exact optimal discovery bound and show that the upper bound is below the discovery or detection boundaries in \citep{CaiSun:2014} and \citep{JJin:2009} (See Figure \ref{fig:f2}).}

For simplicity, from now on, we assume $\boldsymbol{\mu}_1=\mathbf{0}$, $\mu_{2j} =\sqrt{2r \log p}$ for all $j\in S_{01}$ and the features $\mathbf{X}_{\cdot j}$ have the same marginal variance $\sigma^2$, $j=1,\cdots, p$. We also assume that $0<\gamma<\beta<1$, that is $p^{-\gamma}> p^{-\beta}$, which is equivalent to say that on average, the number of marginally non-informative signals correlated with some marginally informative signals is larger than the number of marginally informative signals.  In most cases, the assumption of $\gamma<\beta$ is a reasonable one to make. For example, in genomic data, MUJI markers are usually less sparse than the marginally informative ones.

\begin{theorem}\label{prop:prop6}
\textbf{(Discovery Boundary)} Denote by $d^{\cis}(\beta)$ the discovery boundary associated with the CIS decision rule. Let $\pi=\gamma/\beta$, $0<\pi<1$, then
\begin{itemize}[leftmargin=*]
\item[] (T6.1) For $\sigma=1$, $d^{\cis}(\beta) \leq (1-\sqrt{1-\pi\beta})^2$;
\item[] (T6.2) For $0<\sigma<1$, $d^{\cis}(\beta) \leq \left\{
                          \begin{array}{ll}
                            (1-\sigma\sqrt{1-\pi\beta})^2, & \hbox{ if } 1-\sigma^2 <\beta<1 \\
                            (1-\sigma^2)\pi\beta, & \hbox{ if } 0 < \beta\leq 1-\sigma^2;
                          \end{array}
                        \right.$
\item[] (T6.3) For $\sigma>1$, $d^{\cis}(\beta) \leq \left\{
                              \begin{array}{ll}
                               (1-\sigma\sqrt{1-\pi\beta})^2, & \hbox{ if } 1-1/\sigma^2 <\beta<1 \\
                                0, & \hbox{ if }  0 < \beta <1-1/\sigma^2.
                              \end{array}
                            \right.$
\end{itemize}
\end{theorem}
Define the power of the discovery boundary of some classification procedure as the ratio of the discoverable area to the non-discoverable one. From Theorem \ref{prop:prop6}, we can observe that for any single signal, the discovery boundary from the CIS procedure is at least as powerful as the discovery boundary in \citep{CaiSun:2014}. Moreover, as the MUJI signals get less sparse, the CIS discovery boundary {has more power}. Figure \ref{fig:f2} depicts the discovery boundaries for the optimal screening procedure in \citet{CaiSun:2014} and the CIS procedure with different values of $\pi$. Theorem \ref{prop:prop6} implies that when $3/4 \leq \beta < 1$, the detection boundary from the CIS procedure {has more power than that in \citep{JJin:2009}.}

\begin{figure}[htb!]\label{fig:f2}
\centering
{
  \includegraphics[bb=0.1 13 480 450, clip, scale=0.6]{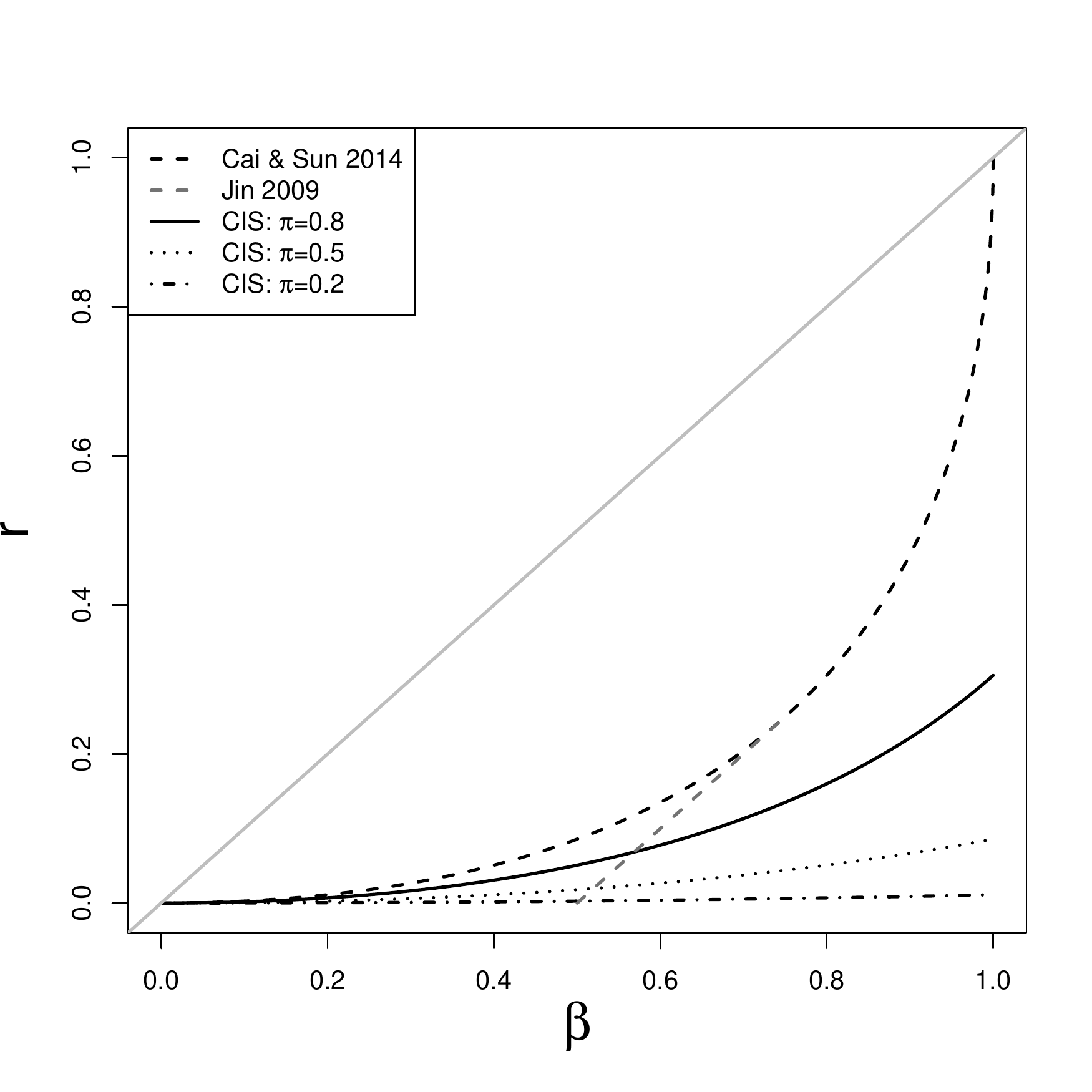}
  \caption{Discovery boundaries for different screening and classification procedures in the case when $\sigma^2=1$. The area above each discovery boundary curve $d(\beta)$ in the $\beta$-$r$ plane is discoverable region with respect to the corresponding procedure and the area below each curve is non-discoverable region. {Where $\sigma^2$ is the marginal variance of a feature and $\pi=\gamma/\beta$ is the ratio of the order for MUJI signals over the order for marginally informative signals, in the scale of $\log p$.}} \label{fig:DB}
  }
\end{figure}

\section{Numerical studies}

{In this section, we assess the finite sample performance of the proposed CIS screening and  classification method  and compare it with the marginal screening method \citep{CaiSun:2014, JJin:2009}.  First, we generate  3 classes  with equal sample size of  $n=150$ with $p=10000$ features, {of which only the first 20 variables are informative}.
}

{The mean structure of the first $20$ variables is given in Table \ref{Tab:means1}. For example, variables 1-4 and 11-14 have mean 0 in both classes 1 and 2, variables 5 and 15 have mean -0.5 in class 1 and mean 2 in class 2, and variables 6-10 and 16-20 have mean 1.5 in class 1 and mean -1.5 in class 2. {Therefore variables 5, 15,  6-10, 16-20 all  are marginally informative for differentiating classes 1 and 2. Variables 1-4 and 11-14,  though  marginally non-informative to differentiate classes 1 and 2,   are correlated with variables 5 and 15 respectively.  Effectively, all the first 20 variables are informative for differentiating classes 1 and 2.}

Variables 1-5 and 11-15 are marginally informative for differentiating classes 2 and 3. Variables 6-10 and 16-20 are marginally non-informative for differentiating classes 2 and 3,  and are set to be independent of variables 1-5 and 11-15 respectively. Therefore, variables 6-10 and 16-20 are non-informative for differentiating classes 2 and 3.

\begin{table}[htb!]
\centering
\caption{Mean structures of the classes in the simulated data}\label{Tab:means1}
\begin{tabular}{lccc}
\hline
Variables& Class 1 &  Class 2 & Class 3\\
 \hline
   {$X_1-X_4$, $X_{11} -X_{14}$} & 0 & 0 & -2.5\\
    $X_5$,  $X_{15}$ & -0.5 & 2 & -2.5\\
   $X_{6}-X_{10}$,  $X_{16}-X_{20}$& 1.5 & -1.5 & -1.5\\
  \hline
\end{tabular}
\end{table}

{The remaining 9,980 variables are independently and identically distributed from $N(0, 1)$ for all classes and are independent of the first 20 variables, hence they are non-informative for differentiating neither classes 1 and 2 nor classes 2 and 3.}

{Our goal is to examine whether  our proposed CIS-classification method renders more power to detect the MUJI variables 1-4 and 11-14 in classifying classes 1 and 2, and whether it performs similarly to the marginal screening methods in classifying classes 2 and 3.}

Regarding the correlation structure between the MUJI variables and the marginally informative variables, we divide the first 20 variables into four correlation blocks $X_1-X_5$, $X_6-X_{10}$, $X_{11}-X_{15}$ and $X_{16}-X_{20}$, and set variables from different correlation blocks  to be independent. Variables within the same block are governed by the same correlation structure. We specifically consider two correlation structures:

\noindent{\emph{Example 1 (Compound Symmetry)}}: each block has a compound symmetry (CS) correlation structure with a correlation coefficient $0.5$. The MUJI variables 1-4 and 11-14 for differentiating classes 1 and 2 are equally correlated with the marginally informative variables 5 and 15, respectively.}

\noindent\emph{Example 2 (First Order Auto-Correlation)}}: each block has a first order auto-correlation (AR1) structure with a correlation coefficient $0.5$. The farther the MUJI variables are away from the marginally informative ones, the weaker their correlations are.

{We generate 500 data sets, each of which  is divided into a  training set  with 100 samples per class and a testing  set with 50 samples per class.
Features are selected  by applying  screening procedures (steps 1-5) on training dataset, described in the previous section. Post-screening is performed on the testing dataset. A thresholding tuning parameter $\tau$ is obtained by a 5-fold cross validation. We also explore the use of  different correlation thresholding values of $\alpha$ and report the corresponding screening results. In both examples, we fix the depth parameter as $m=10$, which controls the {maximum number of edges that connect any variable to a marginally informative variable in its connected component.} In our numerical experience, we have found that the choice of $m$ is not critical.}


We compare the performance of  the proposed method and marginal screening method using the  false positive (FP), false negative (FN), sensitivity (se), specificity (sp),  and the minimum model size  (MMS) {that} is  the  minimum number of features needed to include all informative features. {To assess the  classification performance, we report the   percentage of samples that are misclassified (ER).}
Tables \ref{tab:simResultsCS} and \ref{tab:simResultsAR1} summarize the results
under the CS and AR1 correlation structures.
The presented values  are the averages of the 500 replicates,  except that the median is reported for the MMS.
Tables \ref{tab:simResultsCS} and \ref{tab:simResultsAR1} reveal that in classifying between classes 1 and 2, where MUJI signals exist, the proposed method  provides much smaller false negatives and much {larger sensitivity values. } The smaller MMS for the proposed method also indicates that it identifies MUJI signals more efficiently, compared to the marginal screening method. {When the true correlation is $0.5$, a smaller correlation matrix thresholding value ($\alpha$) gives a better screening result as it includes more marginally weak features, but at the cost of increment of computational time and memory to detect their connected components. In classification between classes 2 and 3, where no MUJI signals exist, the proposed method performs similarly to the marginal screening method in both screening and classification.}

We further investigate the effect of the ``depth" parameter $m$ on the feature selection and classification in \emph{Example 3}, where data are simulated from the CS correlation structure and other settings remain the same as \emph{Example 1}. We fix $\alpha$ at $0.5$.  Table \ref{tab:simResultsdepth} shows that CIS gives satisfactory screening and classification performance even when the depth is  as small  as $m=3$.

\begin{table}[htb!]
\centering \caption{Example 1 (compound symmetry) results}\label{tab:simResultsCS}
\begin{tabular}{lrccccccc}
\hline\hline
&Method & class pair & FP\#&  FN\#& se  & sp & MMS & ER\\
  \hline
   &CIS & 1-2 & 1.1 & 0.1 & 0.994  & 0.999 & 20 & 1.4\\
     &  &  & (3.4) & (0.5) & (0.02) & ($<10^{-3}$) & (0) & (1.1)\\
       && 2-3 & 9.3 & 0 & 1 & 0.999 & 19 & 3.1\\
    $\alpha=0.2$ &     &  & (3.2) & (0) & (0) & ($<10^{-3}$) & (3) & (2.9)\\
   &MS  & 1-2 & 16.4 & 9.1 & 0.54 & 0.998 & 8685 & 3.1\\
     &  &  & (23.9) & (3.2) & (0.2) & (0.002) & (2049) & (4.2)\\
       && 2-3 & 3.0 & 0 & 1 & 0.999 & 10 & 3.4 \\
       && & (14.9) & (0) & (0) & (0.001) & (0) & (0.9)\\
 \hline
  & CIS & 1-2 & 11.8 &  2.5 & 0.88 & 0.999 & 27 & 2.7 \\
    &&  & (22.4) & (3.4) & (0.07) & (0.002) & (7) & (1.5) \\
      & & 2-3 & 5.6 & 0 & 1 & 0.999 & 10 & 2.9\\
     $\alpha=0.5$ &&      & (11.0) & (0) & (0) & ($0.001$) & (10) & (2.1)\\
  & MS  & 1-2 & 18.0 & 9.0 & 0.55 & 0.998 & 8634 & 3.0 \\
    & &  & (29.1) & (3.1) & (0.15) & (0.003) & (2060) & (4.1) \\
      & & 2-3 & 3.5 & 0 & 1 & 0.999 & 10 & 3.4\\
       &&  & (17.1) & (0) & (0) & (0.002) & (0) & (1.0)\\
 \hline
   &CIS & 1-2 & 12.2 & 7.8 & 0.61 & 0.999 & 21 & 1.6\\
    &&  & (19.6) & (0.5) & (0.02) & (0.002) & (2) & (0.9) \\
     &  & 2-3 & 2.7 & 0 & 1 & 0.999 & 10 & 3.4 \\
   $\alpha=0.9$&  &  & (9.5) & (0) & (0) & ($<10^{-3}$) & (0) & (0.9)\\
  & MS  & 1-2 & 17.3 & 9.1 & 0.54 & 0.998 &  8876 &3.1 \\
    &&  & (25.8) & (3.2) & (0.2) & (0.003) & (1913) & (4.3) \\
      & & 2-3 & 2.7 & 0 & 1 & 0.999 & 10 & 3.4 \\
    &&  & (9.5) & (0) & (0) & ($<10^{-3}$) & (0) & (0.9)\\
  \hline\hline
 \multicolumn{9}{l}{\parbox{12cm}{\vspace{0.1in}
\scriptsize{* MS=Marginal screening ignoring the inter-feature correlation.\\
* FP=false positive, FN= false negative, se=sensitivity, sp=specificity, MMS=median of model sizes based on 500 replicated experiments, ER=misclassification error rate (percentage of samples whose class label were mistakenly predicted). Numbers in parentheses are mean standard errors for FP, FN, se, sp, ER and IQRs for MMS.\\
}}} \normalsize
\end{tabular}
\end{table}

\begin{table}[htb!]
\centering \caption{Example 2 (AR1) results}\label{tab:simResultsAR1}
\begin{tabular}{lrccccccc}
\hline\hline
&Method & class pair & FP\#&  FN\# & se  & sp & MMS & ER\\
\hline
  & CIS & 1-2 & 0.9 & 0.05 & 0.997  & 0.999 & 20 & 0.2\\
     &  &  & (3.2) & (0.3) & (0.01) & ($<10^{-3}$) & (0) & (0.9)\\
     &  & 2-3 & 9.9 & 0 & 1 & 0.999 & 17 & 0.9\\
   $\alpha=0.2$    &    &  & (4.7) & (0) & (0) & ($<10^{-3}$) & (5) & (1.5)\\
   &MS  & 1-2 & 19.9 & 9.6 & 0.52 & 0.998 & 8774 & 2.1\\
     &  &  & (35.4) & (3.6) & (0.2) & (0.003) & (1998) & (5.0)\\
       && 2-3 & 4.5 & 0 & 1 & 0.999 & 10 & 1.2 \\
       && & (20.5) & (0) & (0) & 0.002 & (0) & (0.7)\\
 \hline
  & CIS & 1-2 & 9.1 &  5.3 & 0.74 & 0.999 & 24 & 0.6 \\
   & &  & (15.5) & (3.4) & (0.17) & (0.002) & (5) & (1.3) \\
     &  & 2-3 & 10.9 & 0 & 1 & 0.999 & 20 & 1.4\\
    $\alpha=0.5$ &       &  & (23.2) & (0) & (0) & (0.002) & (19) & (1.7)\\
   &MS  & 1-2 & 11.2 & 11.0 & 0.45 & 0.999 & 9354 & 3.3 \\
     &&  & (16.1) & (4.7) & (0.23) & (0.002) & (418) & (6.1) \\
       && 2-3 & 4.6 & 0 & 1 & 0.999 & 10 & 1.4\\
       &&  & (21.7) & (0) & (0) & (0.002) & (0) & (0.7)\\
 \hline
   &CIS & 1-2 & 1.3 & 9.1 & 0.54 & 0.999 & 20 & 0.4\\
    &&  & (3.4) & (1.9) & (0.09) & ($<10^{-3}$) & (0) & (1.2)\\
      & & 2-3 & 0.2 & 0.008 & 0.999 & 0.999 & 10 & 0.9 \\
      $\alpha=0.9$  & & & (1.3) & (0.09) & (0.009) & (0.001) & (0) & (0.6)\\
   &MS  & 1-2 & 1.9 & 11.7 & 0.41 & 0.999 & 7470 & 5.6\\
    &&  & (4.3) & (5.3) & (0.3) & ($<10^{-3}$) & (3679) & (7.4)\\
     &  & 2-3 & 0.2 & 0.02 & 0.998 & 0.999 & 10 & 1.0\\
    &&  & (1.3) & (0.4) & (0.04) & ($<10^{-3}$) & (0) & (0.7)\\
  \hline\hline
\end{tabular}
\end{table}

\begin{table}[htb!]
\centering \caption{Example 3 results}\label{tab:simResultsdepth}
\begin{tabular}{lrccccccc}
\hline\hline
&Method & class pair & FP\#&  FN\# & se  & sp & MMS & ER\\
\hline
  & CIS & 1-2 & 11.8 &  2.5 & 0.88 & 0.999 & 27 & 2.7 \\
    &&  & (22.4) & (3.4) & (0.07) & (0.002) & (7) & (1.5) \\
      & & 2-3 & 5.6 & 0 & 1 & 0.999 & 10 & 2.9\\
  $m=10$   &   &  & (11.0) & (0) & (0) & ($0.001$) & (10) & (2.1)\\
   &MS  & 1-2 & 18.0 & 9.0 & 0.55 & 0.998 & 8634 & 3.0 \\
     &&  & (29.1) & (3.1) & (0.15) & (0.003) & (2060) & (4.1) \\
       && 2-3 & 3.5 & 0 & 1 & 0.999 & 10 & 3.4\\
       &&  & (17.1) & (0) & (0) & (0.002) & (0) & (1.0)\\
\hline
   &CIS & 1-2 & 12.7 & 4.1 & 0.80  & 0.999 & 28 & 2.1\\
     &  &  & (19.9) & (2.9) & (0.14) & (0.002) & (8) & (1.8)\\
       && 2-3 & 8.6 & 0 & 1 & 0.999 & 10 & 4.3\\
$m=3$ &      &  & (20.0) & (0) & (0) & (0.002) & (8) & (1.7)\\
   &MS  & 1-2 & 21.6 & 9.3 & 0.54 & 0.998 & 8752 & 2.6 \\
     &  &  & (33.5) & (3.4) & (0.2) & (0.003) & (1749) & (4.3)\\
      & & 2-3 & 1.6 & 0 & 1 & 0.999 & 10 & 2.8 \\
       && & (6.4) & (0) & (0) & ($<10^{-3}$) & (0) & (0.7)\\
   \hline
   &CIS & 1-2 & 15.1 & 5.9 & 0.71 & 0.998 & 24 & 1.5\\
    &&  & (24.3) & (1.8) & (0.09) & (0.002) & (10) & (1.6)\\
       && 2-3 & 7.3 & 0 & 1 & 0.999 & 10 & 3.8 \\
 $m=1$ &  &  & (20.1) & (0) & (0) & (0.002) & (7) & (1.5)\\
  & MS  & 1-2 & 21.1 & 9.6 & 0.52 & 0.998 & 8772 & 2.9\\
    &&  & (32.8) & (3.7) & (0.2) & (0.003) & (1630) & (4.6)\\
      & & 2-3 & 1.4 & 0 & 1 & 0.999 & 10 & 2.8\\
    &&  & (6.7) & (0) & (0) & ($<10^{-3}$) & (0) & (0.7)\\
  \hline\hline
\end{tabular}
\end{table}

\section{Real data analysis}

There are 4 functional types of tissues among the 62 kidney tissue samples in the kidney transplant study \citep{FLENCHER:2004}: 17 normal donor kidneys (C), 19 well-functioning kidneys more than 1-year post-transplant (TX), 13 biopsy-confirmed  acute rejection (AR), and 13 acute dysfunction with no rejection (NR).  Each sample has gene profile assayed from kidney biopsies and peripheral blood lymphocytes on 12,625 genes. Distinguishing these 4 types of tissues is important in balancing the need for immunosuppression to prevent rejection and minimizing drug-induced toxicities.

We assess the classification performance by predicting each individual's class using the leave-one-out procedure. Specifically, for each sample, we classify it between any class pair using the rest samples belong to the class pair to select informative features and optimal tuning parameters. The thresholding parameter $\tau$ is chosen by 5-fold cross validation. As the median absolute value of the correlation coefficient between two features for our data is $0.25$, the covariance (or the correlation as we have standardized each feature) thresholding parameter $\alpha$ is fixed at $0.2$. The depth  parameter is set as $m=10$. {Since there are 4 classes, a majority voting strategy is then employed to decide which class the sample is eventually classified to. For example, if a sample is classified to class C between the classes C and TX, to C between C and AR, to C between C and NR, to TX between TX and AR, to NR between TX and NR, and to NR between AR and NR, then the sample is eventually classified to C. The pairwise comparison strategy which reduces the multi-class problem to multiple two-class problems \citep{Tax_usingtwo-class:2002}, and the leave-one-out plus majority voting classification error is 6 out of 62 samples.}

In terms of selecting informative variables, we draw 100 bootstrap samples for each pairwise two-class comparison and use 5-fold cross-validation to select the tuning parameters. Top genes with high selection frequencies for each class pair are identified and
reported  in Table \ref{tab:realData}.  MUJI genes are indicated in the third column of Table \ref{tab:realData}.
The last column lists literature  that validate the biological relevance of the selected genes.

Our method identifies some MUJI genes as well as some marginally informative genes supported by the clinical literature. For example,  \emph{DDX19A} has been identified as a MWJI gene which has high discrimination power between classes C and AR. \citep{JLI_DDX19A:2015} identified \emph{DDX19A}, a member of the DEAD/H-box protein family, as a novel component of \emph{NLRP3} inflammasome. \textit{NLRP3} inflammasome plays a major role in innate immune and inflammatory pathologic responses, which is relevant to post-renal-transplantation recovery. To our knowledge, \emph{DDX19A} has not been reported elsewhere as a statistically informative gene for post-renal-transplantation rejection types.

\begin{figure}
  \centering
  \includegraphics[bb=20 10 563 210, clip, scale=0.6]{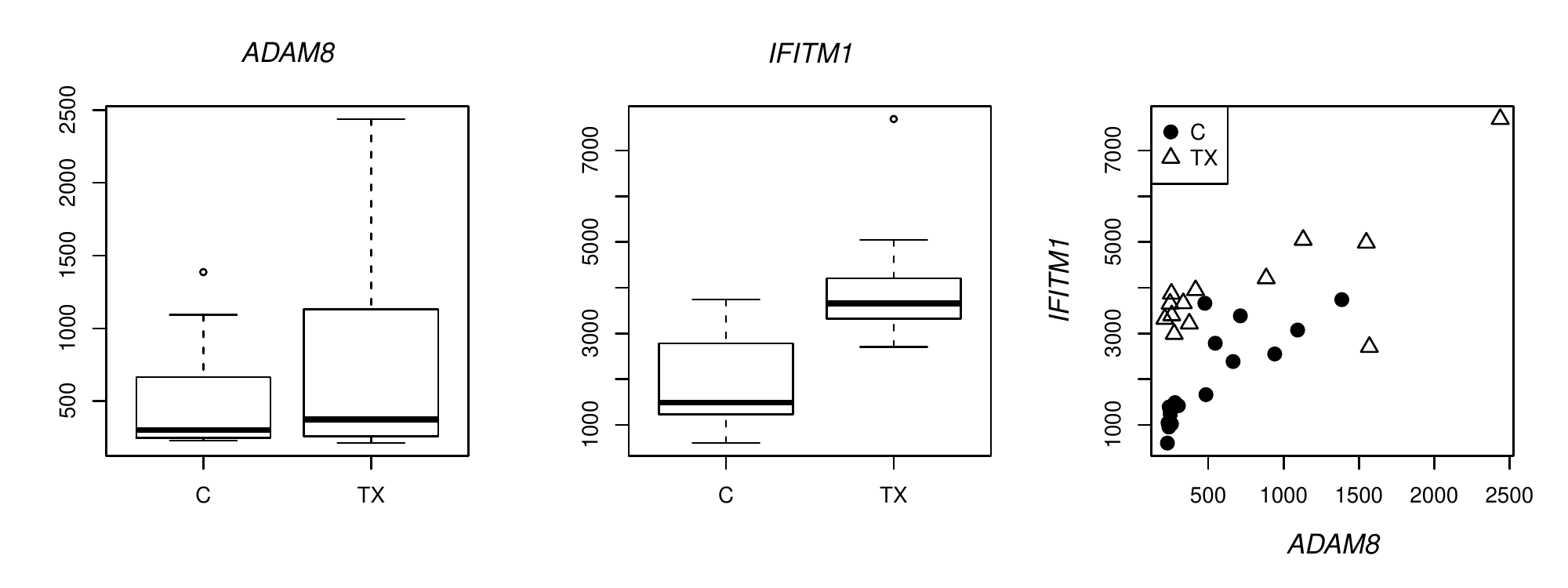}
  \includegraphics[bb=20 10 563 210, clip, scale=0.6]{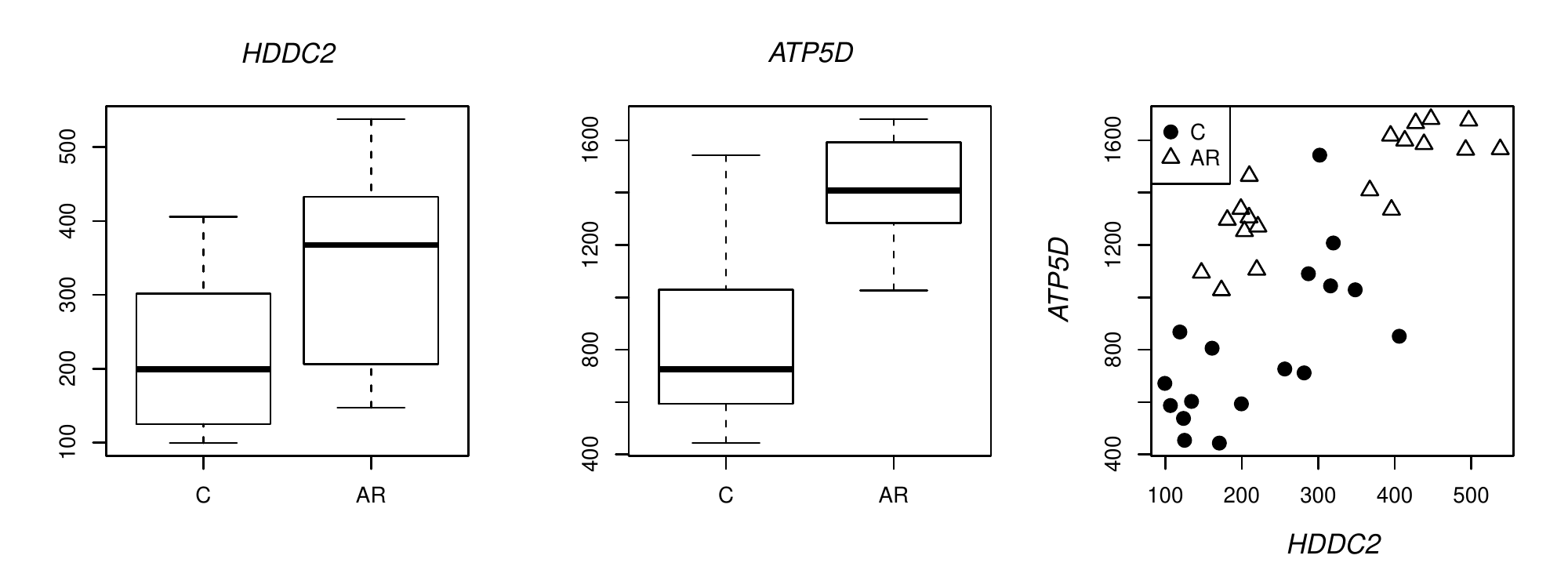}
  \includegraphics[bb=20 10 563 210, clip, scale=0.6]{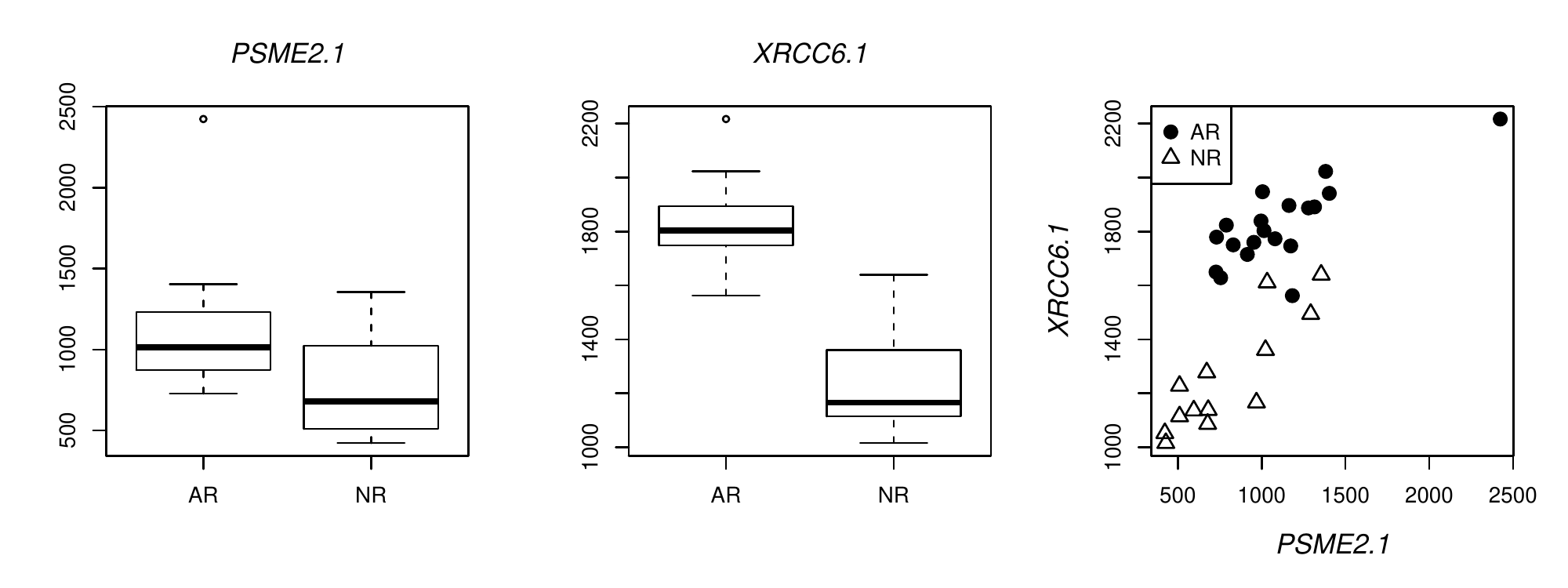}
  \caption{Marginal and joint effects of MUJI genes.} \label{fig:realData}
\end{figure}

The individual and joint effects for some MUJI genes are illustrated in Figure \ref{fig:realData}. The marginal effects of those MUJI genes and their correlated marginally discriminative genes, together with rank of importance scores of the MUJI genes are summarized in Table \ref{tab:mujiGeneFigure}. For example, in the top panel of Figure \ref{fig:realData}, the marginal standardized mean difference between C and TX for the gene \emph{ADAM8} is ranks 5,071 out of 12,625 genes. While gene \emph{ADAM8} is highly correlated with a marginally discriminative gene \emph{IFITM1}, whose standardized marginal mean difference ranks 22. By incorporating the covariance structure, the CIS procedure promote rank of the IS for gene \emph{ADAM8} to 18.

\begin{table}[htb!]
\centering \caption{Top selected genes from the kidney transplant data}\label{tab:realData}
\begin{tabular}{lccl}
\hline\hline
Gene & class pair & \ \ MUJI$^{*}$? \ \ & literature \\
 & (selection frequency \%$^{**}$) & & evidence \\
 \hline
     & & &\\
   \emph{PABPC4} & C-TX(54) & $\boldsymbol\surd$ &\scriptsize{\citet{FFChen_PABPC:2010}}\\
   \emph{GNB2L1} & C-AR(66) & $\mathbf{X}$ &\scriptsize{\citet{WangS:2003}}\\
     & & &\scriptsize{\citet{TCGARN:2013}}\\
     & & &\scriptsize{\citet{LLiCancerCell:2013}}\\
   \emph{TG(Thyroglobulin)} &C-AR(57)& $\boldsymbol\surd$& \scriptsize{\citet{WJu:2009}}\\
     & & &\scriptsize{\citet{HWu_TG:2009}}\\
     & & &\scriptsize{\citet{Sellittia_TG:2000}}\\
     & & &\scriptsize{\citet{YLuo_TG:2014}}\\
   \emph{PTTG1} & C-AR(52) & $\boldsymbol\surd$&\scriptsize{\citet{Hamid_PTTG1:2005}}\\
     & & &\scriptsize{\citet{Wondergem_PTTG1:2012}}\\
     & & &\scriptsize{\citet{CWei_PTTG1:2015}}\\
     & & &\scriptsize{\citet{Kristopher_PTTG1:2003}}\\
   \emph{DDX19A} & C-AR(44) & $\boldsymbol\surd$ &\scriptsize{\citet{JLI_DDX19A:2015}}\\
   \emph{ASH2L} & C-AR(40) & $\boldsymbol\surd$ &\scriptsize{\citet{DeSanta_ASH2L:2007}}\\
   \emph{MAPK3}  & C-ADNR(100) & $\boldsymbol\surd$ &\scriptsize{\citet{SLZhang_MAPK:1999}} \\
     & & &\scriptsize{\citet{Cassidy_MAPK:2012}}\\
     & & &\scriptsize{\citet{Awazu_MAPK:2002}}\\
     & & &\scriptsize{\citet{SLZhang_MAPK:2000}}\\
     & & &\scriptsize{\citet{SLZhang_MAPK:2002}}\\
     & & &\scriptsize{\citet{EKKim_MAPK:2010}}\\
   \emph{TMEM199}  & AR-ADNR(53) & $\boldsymbol\surd$ &\scriptsize{\citet{MCHogan_TMEM:2003}} \\
     & & &\scriptsize{\citet{MCHogan_TMEM:2003}}\\
   \emph{KIAA1467}  & AR-ADNR(40) & $\boldsymbol\surd$ &\scriptsize{\citet{WSui_KIAA1467:2013}} \\
   \ \ \ ... & ... & ... &\ \ \ ... \\
        & & &\\
  \hline\hline
 \multicolumn{4}{l}{\parbox{12cm}{\vspace{0.1in}
\scriptsize{
*  {MUJI= a marginally uninformative but jointly informative.}\\
** selection frequency= the percentage of gene selected out of $100$ bootstrapped data sets.
}}} \normalsize
\end{tabular}
\end{table}

\begin{table}
\centering \caption{Marginal and joint effects of MUJI genes}\label{tab:mujiGeneFigure}
\begin{tabular}{lccccccc}
\hline\hline
MUJI  & $\hat{\delta}_j^*$ of & rank of & correleted & $\hat{\delta}_j$ of & rank of&correlation  & rank of \\
gene & MUJI& MUJI & MI$^{**}$ & MI & MI& & $IS_j$\\
& gene & gene & gene & gene & gene& & \\
 \hline
\emph{ADAM8} & -0.48  & {5071} & \emph{IFITM1} & -1.35 & 22 & 0.70 & {18}\\
\emph{HDDC2} & -0.82  & {2497} & \emph{ATP5D} & -1.53 & 51 & 0.71 & {77}\\
\emph{PSME2.1} & 0.84  & {3962} & \emph{XRCC6.1} & 1.71 & 22 & 0.74 & {22}\\
  \hline\hline
 \multicolumn{8}{l}{\parbox{12cm}{\vspace{0.1in}
\scriptsize{
*  $\hat{\delta}_j$ = standardized meanginal mean difference of gene j.\ \ ** MI= marginally informative
}}} \normalsize

\end{tabular}
\end{table}

The following Figure \ref{fig:MUJIeffect} illustrates the classification effect of a MUJI signal, gene \emph{IPO5}, when considered with a marginally stronger signal, gene \emph{TTC37}. Considered by itself, the minimal misclassification number of gene \emph{TTC37} is 10, as showed in the left panel of Figure \ref{fig:MUJIeffect}. While jointly considered with gene \emph{IPO5}, the minimal misclassification number reduces to 2, as showed in the right panel of Figure \ref{fig:MUJIeffect}. Figure \ref{ffig:MUJIeffect} clearly demonstrates more discriminating power  when two genes are considered jointly than when considered individually.

\begin{figure}[htb!]\label{fig:MUJIeffect}
  \centering
  \includegraphics[scale=0.5]{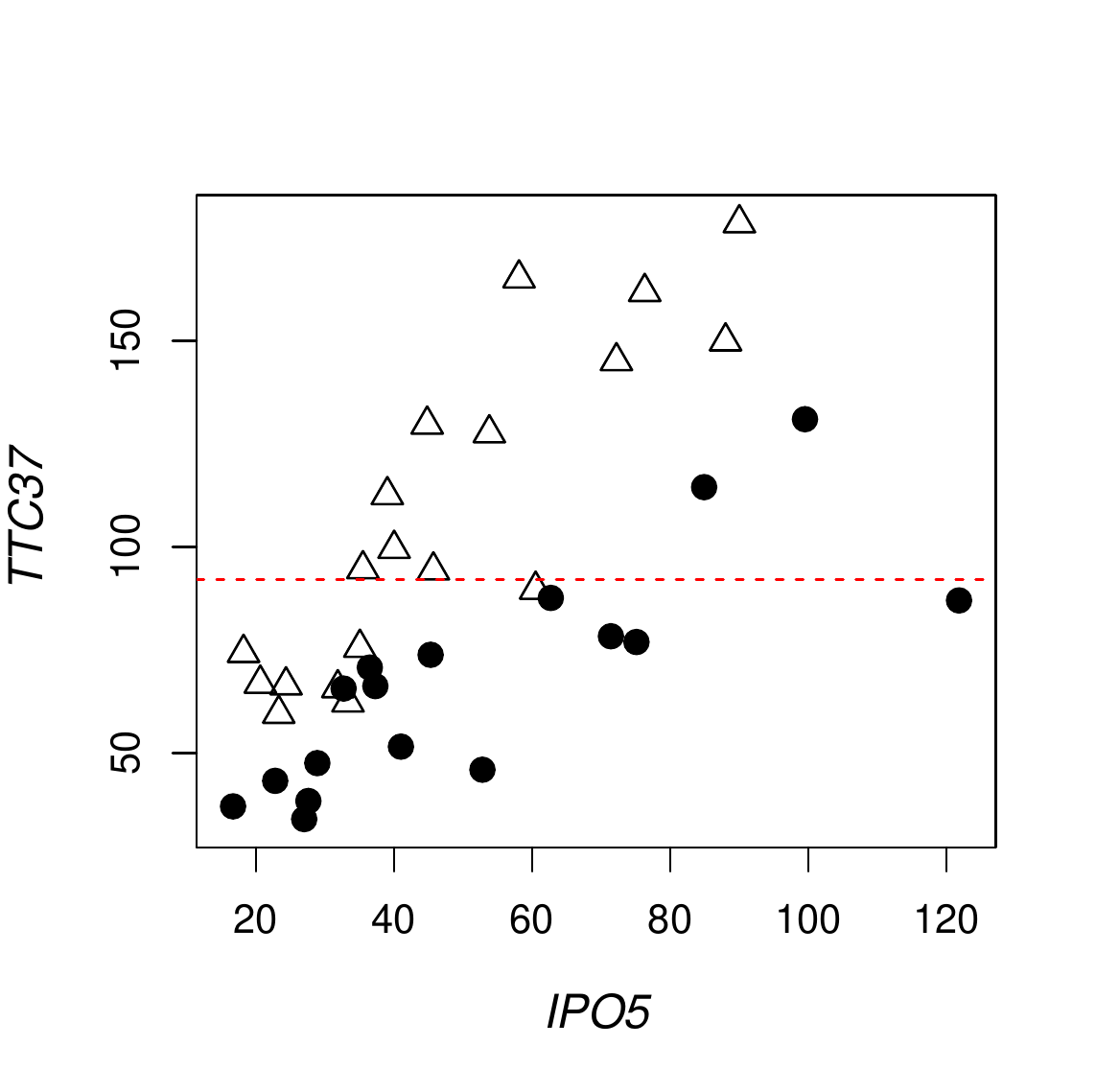}
  \includegraphics[scale=0.5]{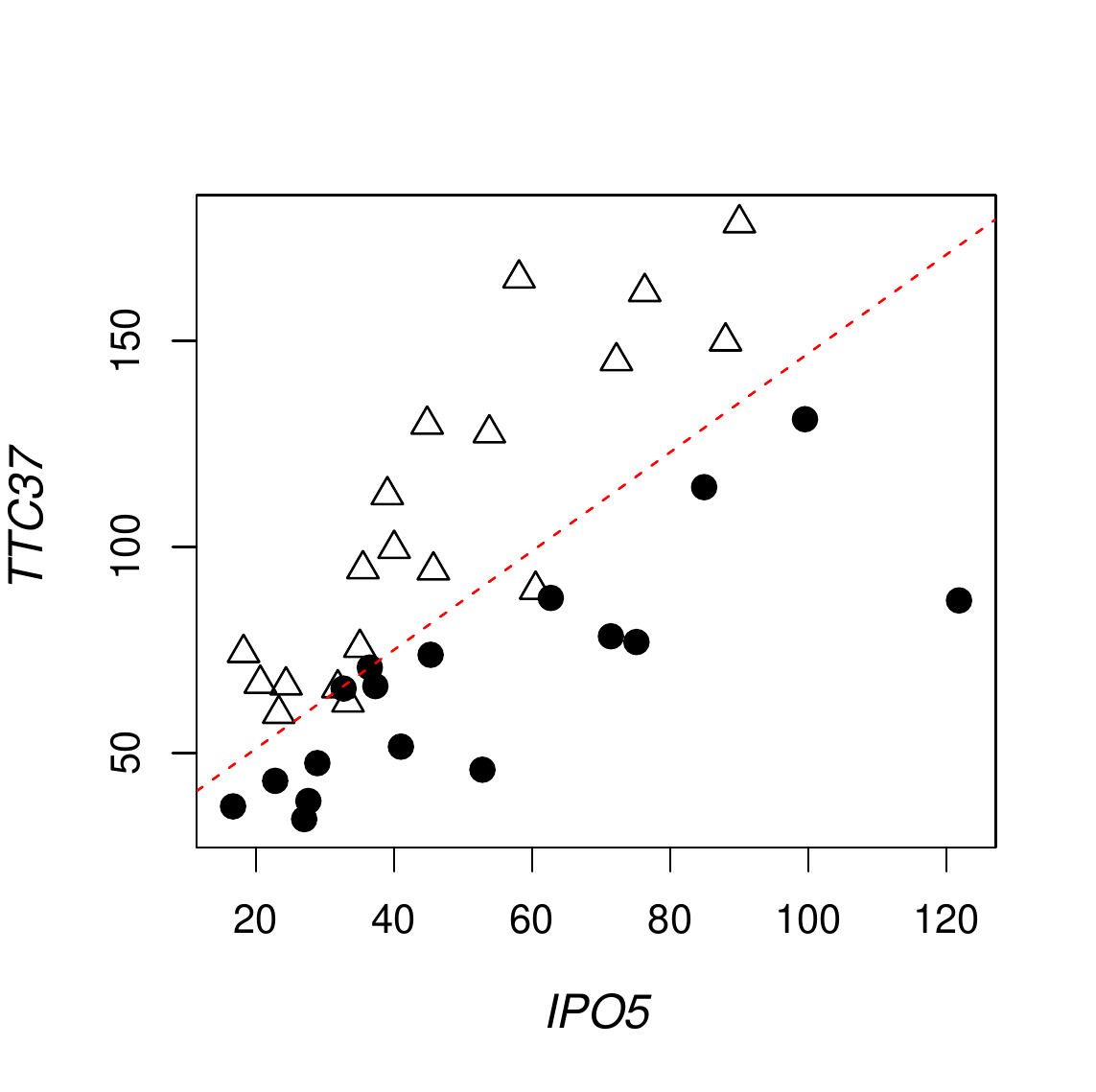}
\caption{Illustration of effect of an MUJI signal on classification.}
\end{figure}

\section{Discussion}
{Leveraging dependence among features}, we propose a covariance-insured screening and classification procedure for weak signal detection and ultrahigh-dimensional classification.
 The procedure is accompanied by  easily implementable algorithms and renders nice theoretical properties, such as model selection consistency with optimal misclassification rates.

The proposed method has a wide range of applications, such as classification of  disease subtypes based on spatially correlated fMRI brain image data and classification of portfolio categories based on time series data for prices. The local correlation structure can be parameterized with fewer  parameters (such as AR1) to further improve the estimation accuracy and efficiency of precision matrices.

Several aspects of the proposed method can inspire future work in several directions. {A key component of our approach is the notion of identifying block-diagonal precision matrices by thresholding the sample covariance matrices. On the other hand, intensive  biomedical  research  has  generated  a  large  body
of biological knowledge, e.g. pathway information,  that may help specify the structure of such matrics. How to effectively utilize such information, and  further balance priori knowledge and data-driven approaches to achieve optimal decision rules remains rather challenging.}

We assume a common covariance matrix across classes. However this restrictive assumption can be relaxed by allowing heteroscedastic normal models, which are more realistic in many  practical applications.  As stated in the proofs, the normality assumption can also be relaxed, and be replaced by  sub-Gaussian families. This will facilitate modeling non-Gaussian data with heavy-tails.

There is much room for development of more efficient and faster algorithms. For instance,  when the whole feature space can be divided into uncorrelated subspaces, such as different chromosomes in the genome or different functional regions in the brain,  parallel computing on multiple partitioned feature spaces may be available. We will pursue this in future work.

\appendix
\section{Proofs of the main results}\label{app}

We first state the Gaussian tail inequality, which will be used throughout in Appendix \ref{app}. Let $Z$ be a standard normal random variable. Then for any $t>1$,
\begin{equation*}
P(Z>t) \leq \frac{1}{\sqrt{2\pi}t}e^{-\frac{t^2}{2}}.
\end{equation*}

Denote by $\tilde{\boldsymbol{\Sigma}}^{\alpha}=\hat{\boldsymbol{\Sigma}}_{ij} \mathbbm{1}(|\hat{\boldsymbol{\Sigma}}_{ij}|\geq \alpha)$ with $\hat{\boldsymbol{\Sigma}}=\frac{1}{n}\sum_{i=1}^n (\mathbf{X}_i-\bar{\mathbf{X}})(\mathbf{X}_i-\bar{\mathbf{X}})'$ being the sample covariance matrix. \citet{Bickel_Levina:2008} showed that for $\alpha=O(\sqrt{\log p/n})$, $\|\hat{\boldsymbol{\Sigma}}^{\alpha} - \boldsymbol{\Sigma}\| = O(C_{h,p}(n^{-1}\log p)^{(1-h)/2})$. Furthermore, \citet{Bickel_Levina:2008} and \citet{Fan_Liao:2011} showed that,
\begin{equation}\label{eqn:precisionBound}
\|(\tilde{\boldsymbol{\Sigma}}^{\alpha})^{-1}- \boldsymbol{\Sigma}^{-1} \| =O\left( C_{h,p}\left(\frac{\log p}{n}\right)^{(1-h)/2}\right).
\end{equation}
Therefore, under the sparsity assumptions, the CIS estimated precision matrix satisfies $\|\hat{\boldsymbol{\Omega}}^{\alpha, \tau}- \boldsymbol{\Omega} \| \leq M_0 C_{h,p}( n^{-1}\log p )^{(1-h)/2}$ for some constant $M_0$ and sufficiently large $n$. The following Lemma \ref{lem:lemE} is useful in controlling the size of non-zero entries in a row of $\hat{\boldsymbol{\Omega}}$. Its proof is similar to that of Theorem 2 in \citep{SLuo:2014} and thus is omitted.


\begin{lem}\label{lem:lemE}
Let $\alpha\equiv\alpha(n,p)=O\left((\log p/n)^{1/2}\right)$. Under assumptions (A2)-(A4), $P\left(|\hat{\mathcal{E}}_{j, \alpha}| \leq O(n^{(1-\xi)})\right) \geq 1- C_1\exp(-C_2n^{\xi})$ for some positive constants $C_1$ and $C_2$.
\end{lem}

\noindent\textbf{Proof of Theorem \ref{prop:FNC}.}
Denote $S^{\scriptsize{CIS}}(\alpha, \tau; \nu_n)$ for short by $S_{\scriptsize{CIS}}^{\alpha, \tau; \nu_n}$ and let $\mathcal{S}=\cup_{g=1}^{k(\alpha, \tau)} S_g$. Proving (\ref{eqn:FNC}) is equivalent to proving the following when $n \rightarrow \infty$
\begin{eqnarray}
&& P(|S_{CIS}^{\alpha, \tau; \nu_n}\cap S_0| > (1-\epsilon)|S_0| ) \label{eqn:prfThm1.1}\\
&=& P\left(\frac{1}{|S_0|}\sum_{j\in S_0} \mathbbm{1}\left(\sum_{j'\in \mathcal{S}} \left|\hat{\Omega}_{jj'}^{\alpha, \tau} (\bar{X}_{\cdot j'}^{(1)}-\bar{X}_{\cdot j'}^{(2)})\right| >\nu_n \right) > 1-\epsilon \right) \rightarrow 0. \nonumber
\end{eqnarray}
Let $T^{\alpha, \tau}_j=\mathbbm{1}\left(\sum_{j'\in \mathcal{S}} |\hat{\Omega}_{jj'}^{\alpha, \tau} (\bar{X}_{\cdot j'}^{(1)} -\bar{X}_{\cdot j'}^{(2)})| > \nu_n \right)$. We first develop an upper bound for $E(T^{\alpha, \tau}_j)$. Notice that when $\hat{\Omega}_{jj'}$ is estimated as zero, it does not contribute to $\sum_{j'\in \mathcal{S}} |\hat{\Omega}_{jj'}^{\alpha, \tau} (\bar{X}_{\cdot j'}^{(1)} -\bar{X}_{\cdot j'}^{(2)})|$. Therefore
\begin{eqnarray*}
E(T^{\alpha, \tau}_j) &=&  P\left(\sum_{j'\in \mathcal{S}} |\hat{\Omega}_{jj'}^{\alpha, \tau} (\bar{X}_{\cdot j'}^{(1)}-\bar{X}_{\cdot j'}^{(2)})| > \nu_n\right)\\
&=& P\left(\sum_{j'\in \hat{\mathcal{E}}_{j, \alpha}} |\hat{\Omega}_{jj'}^{\alpha, \tau} (\bar{X}_{\cdot j'}^{(1)}-\bar{X}_{\cdot j'}^{(2)})| > \nu_n\right)\\
&\leq& P\left(\sum_{j'\in \hat{\mathcal{E}}_{j, \alpha}} |(\hat{\Omega}_{jj'}^{\alpha, \tau} -\Omega_{jj'}) (\bar{X}_{\cdot j'}^{(1)}-\bar{X}_{\cdot j'}^{(2)})| >\nu_n\right) \\
& & +P\left(\sum_{j'\in \hat{\mathcal{E}}_{j, \alpha}} |\Omega_{jj'} (\bar{X}_{\cdot j'}^{(1)}-\bar{X}_{\cdot j'}^{(2)})| >\nu_n\right)
\end{eqnarray*}
By (\ref{eqn:precisionBound}) and the assumption (A1) that $C_{h,p}$ is bounded away from infinity, $\hat{\boldsymbol{\Omega}}^{\alpha, \tau} - \boldsymbol{\Omega}$ is a bounded operator, i.e. $\|\hat{\boldsymbol{\Omega}}^{\alpha, \tau} - \boldsymbol{\Omega}\| \leq M <\infty$, for some $M>0$. Then,
\begin{eqnarray}
&& \ \ \ \ \ \ \ P\left(\sum_{j'\in \hat{\mathcal{E}}_{j, \alpha}} |(\hat{\Omega}_{jj'}^{\alpha, \tau} -\Omega_{jj'}) (\bar{X}_{\cdot j'}^{(1)}-\bar{X}_{\cdot j'}^{(2)})| >\nu_n\right) \label{eqn:tmp2_0}\\
&=& P\left((\bar{\mathbf{X}}^{(1)}_{\cdot\hat{\mathcal{E}}_{j, \alpha}}-\bar{\mathbf{X}}^{(2)}_{\cdot\hat{\mathcal{E}}_{j, \alpha}})'(\hat{\boldsymbol{\Omega}}^{\alpha, \tau} -\boldsymbol{\Omega})'_{j\hat{\mathcal{E}}_{j, \alpha}}(\hat{\boldsymbol{\Omega}}^{\alpha, \tau} -\boldsymbol{\Omega})_{j\hat{\mathcal{E}}_{j, \alpha}} (\bar{\mathbf{X}}^{(1)}_{\cdot\hat{\mathcal{E}}_{j, \alpha}}-\bar{\mathbf{X}}^{(2)}_{\cdot\hat{\mathcal{E}}_{j, \alpha}})>\nu_n^2 \right) \nonumber\\
&\leq& P\left(\lambda_{\mbox{\scriptsize{max}}}\left((\hat{\boldsymbol{\Omega}}^{\alpha, \tau} -\boldsymbol{\Omega})'_{j.}(\hat{\boldsymbol{\Omega}}^{\alpha, \tau} -\boldsymbol{\Omega})_{j.}\right) \|\bar{\mathbf{X}}^{(1)}_{\cdot\hat{\mathcal{E}}_{j, \alpha}}-\bar{\mathbf{X}}^{(2)}_{\cdot\hat{\mathcal{E}}_{j, \alpha}}\|_2^2>\nu_n^2 \right) \nonumber\\
&\leq& P\left(M \|\bar{\mathbf{X}}^{(1)}_{\cdot\hat{\mathcal{E}}_{j, \alpha}}-\bar{\mathbf{X}}^{(2)}_{\cdot\hat{\mathcal{E}}_{j, \alpha}}\|_2^2>\nu_n^2 \right), \nonumber
\end{eqnarray}
where $\bar{\mathbf{X}}^{(1)}_{\cdot\hat{\mathcal{E}}_{j, \alpha}}-\bar{\mathbf{X}}^{(2)}_{\cdot\hat{\mathcal{E}}_{j, \alpha}}$ is a vector of length $|\hat{\mathcal{E}}_{j, \alpha}|$ and $\|\cdot\|_2$ is the $L^2$ norm of a vector. The last step stems from that the maximum eigenvalue of a principal submatrix of a symmetric positive-definite matrix is smaller than or equal to the maximum eigenvalue of the original matrix. Or $\lambda_{\mbox{\scriptsize{max}}}\left((\hat{\boldsymbol{\Omega}}^{\alpha, \tau} -\boldsymbol{\Omega})'_{j\cdot}(\hat{\boldsymbol{\Omega}}^{\alpha, \tau} -\boldsymbol{\Omega})_{j\cdot}\right) \leq \lambda_{\mbox{\scriptsize{max}}}\left((\hat{\boldsymbol{\Omega}}^{\alpha, \tau} -\boldsymbol{\Omega})'(\hat{\boldsymbol{\Omega}}^{\alpha, \tau} -\boldsymbol{\Omega})\right)\leq M$.
By assumption (A4), $\|\boldsymbol{\Omega}\| \leq \kappa_2$. Following the same argument above, we also have
\begin{equation}\label{eqn:tmp2}
P\left(\sum_{j'\in \hat{\mathcal{E}}_{j, \alpha}} |\Omega_{jj'} (\bar{X}^{(1)}_{\cdot j'}- \bar{X}^{(2)}_{\cdot j'})| >\nu_n\right) \leq P\left(\kappa_2 \|\bar{\mathbf{X}}^{(1)}_{\cdot\hat{\mathcal{E}}_{j, \alpha}}-\bar{\mathbf{X}}^{(2)}_{\cdot\hat{\mathcal{E}}_{j, \alpha}}\|_2^2>\nu_n^2 \right).
\end{equation}
Combining (\ref{eqn:tmp2_0}) and (\ref{eqn:tmp2}), we have
\begin{eqnarray*}
E(T^{\alpha, \tau}_j) &\leq& 2 P\left(C \|\bar{\mathbf{X}}^{(1)}_{\cdot\hat{\mathcal{E}}_{j, \alpha}}-\bar{\mathbf{X}}^{(2)}_{\cdot\hat{\mathcal{E}}_{j, \alpha}}\|_2^2>\nu_n^2 \right) \\
&\leq& 2 P\left(C \|\bar{\mathbf{X}}^{(1)}_{\cdot\hat{\mathcal{E}}_{j, \alpha}}-\bar{\mathbf{X}}^{(2)}_{\cdot\hat{\mathcal{E}}_{j, \alpha}}\|_1>\nu_n \right) \nonumber\\
&\leq& 2P\left(|\bar{X}^{(1)}_{\cdot j'}-\bar{X}^{(2)}_{\cdot j'}|> \nu_n/(C |\hat{\mathcal{E}}_{j, \alpha}| ) \right)
\end{eqnarray*}
with some $j' \in \hat{\mathcal{E}}_{j, \alpha}$ and $C=\max\{M, \kappa_2\}$ and $\|\cdot\|_1$ being the $L^1$ norm of a vector. If $j \in S_{01}$, by Lemma \ref{lem:lemE}, with a probability at leat $1- C_1\exp(-C_2n^{\xi})$,
\begin{eqnarray*}
&&P\left(|\bar{X}^{(1)}_{\cdot j'}-\bar{X}^{(2)}_{\cdot j'}|> \nu_n/(C |\hat{\mathcal{E}}_{j, \alpha}| ) \right)\\
&\leq& 2P\left(\bar{X}^{(1)}_{\cdot j'}-\bar{X}^{(2)}_{\cdot j'}-\delta_{j'} > \nu_n/(C |\hat{\mathcal{E}}_{j, \alpha}| ) -\delta_{j'} \right)\nonumber\\
&\leq& 2P\left(Z > n^{(1-\xi)}\sqrt{r \log p}/(C n^{1-\xi})-O(\sqrt{r\log p})\right) \\
&\leq& 2P\left(Z > O(\sqrt{r\log n})\right) \nonumber\\
&\leq& \frac{\sqrt{2}}{\sqrt{\pi r\log n}} n^{-r}.
\end{eqnarray*}
With a similar argument, it is easy to show that the same inequality also holds for $j \notin S_{01}$. As a result $E(T^{\alpha, \tau}_j) = o(n^{-r})$. Note that $ T^{\alpha, \tau}_j$ here are not necessarily independent variables. Using the Hoeffding's inequality for dependent random variables \citep{SAVandeGeer:2002}, we have
\begin{eqnarray*}
&&P\left(\frac{1}{|S_0|}\sum_{j\in S_0} T^{\alpha, \tau}_j > 1-\epsilon \right)\\
&=& P\left(\frac{1}{|S_0|}\sum_{j\in S_0}\left[ T^{\alpha, \tau}_j -E(T^{\alpha, \tau}_j)\right]  > 1-\epsilon-\frac{1}{|S_0|}\sum_{j\in S_0}E(T^{\alpha, \tau}_j)  \right) \nonumber \\
&=& P\left(\frac{1}{|S_0|}\sum_{j\in S_0}\left[ T^{\alpha, \tau}_j -E(T^{\alpha, \tau}_j)\right]  > 1-\epsilon-o(n^{-r})  \right) \nonumber \\
&=& P\left(\frac{1}{\sqrt{|S_0|}}\sum_{j\in S_0}\left[ T^{\alpha, \tau}_j -E(T^{\alpha, \tau}_j) \right] > \sqrt{|S_0|}\left(1-\epsilon-o(n^{-r})\right)\right) \nonumber\\
&\leq& P\left(\sum_{j\in S_0}\left[ \frac{T^{\alpha, \tau}_j}{\sqrt{|S_0|}} -\frac{E(T^{\alpha, \tau}_j)}{\sqrt{|S_0|}} \right] > O(n^{(1-\beta)/2})(1-\epsilon-o(n^{-r}))  \right) \nonumber\\
&\leq& \exp\left\{ \frac{-2O(n^{1-\beta}) (2-2\epsilon-o(n^{-r}))}{\sum_{j\in S_0}\left(\frac{1}{\sqrt{|S_0|}}\right)^2}\right\}\rightarrow 0 \nonumber
\end{eqnarray*}
as  $n\rightarrow \infty$ and $p\rightarrow \infty$. The second last step stems from that $|S_{0}| \geq |S_{01}| = O(p^{1-\beta}) \geq O(n^{1-\beta})$ by assumption. \qed

\noindent\textbf{Proof of Theorem \ref{prop:FPC}.}
Since $E(T^{\alpha, \tau}_j) = o(n^{-r})$, $\zeta_n^{-1}-E(T^{\alpha, \tau}_j) = n^{-r}(1+o(1))$ for each $j\in S_0^c$. Also notice that for each of the $p^{1-\beta}$ marginally informative features, there are at most $p^{1-\gamma}$ features associated with it. Therefore $|S_0| \leq p^{2-\beta-\gamma}$ and $|S^c_0| \geq p(1-p^{1-\beta-\gamma}) \geq n(1-p^{1-\beta-\gamma})$, then
\begin{eqnarray*}
&& P(\frac{|S^{\alpha, \tau; \nu_n}_{CIS}\cap S^c_0|}{|S^c_0|} > \zeta_n^{-1} ) \\
&=& P\left(\frac{1}{|S^c_0|}\sum_{j\in S_0^c} \mathbbm{1}\left(\sum_{j'\in \mathcal{S}} \hat{\Omega}_{jj'}^{\alpha, \tau} (\bar{X}_{1j'}-\bar{X}_{2j'}) >\nu_n \right) > \zeta_n^{-1} \right) \nonumber\\
&=& P\left(\frac{1}{|S^c_0|}\sum_{j\in S_0^c} T^{\alpha, \tau}_j > \zeta_n^{-1}\right) \\
&=& P\left(\frac{1}{|S^c_0|}\sum_{j\in S_0^c}\left[ T^{\alpha, \tau}_j -E(T^{\alpha, \tau}_j) \right] > \zeta_n^{-1}- \frac{1}{|S^c_0|}\sum_{j\in S_0^c}E(T^{\alpha, \tau}_j)\right) \nonumber \\
&=& P\left(\frac{1}{|S^c_0|}\sum_{j\in S_0^c}\left[ T^{\alpha, \tau}_j -E(T^{\alpha, \tau}_j) \right] > n^{-r}(1+o(1))\right) \nonumber\\
&=& P\left(\sum_{j\in S_0^c}\left[ \frac{T^{\alpha, \tau}_j}{\sqrt{|S^c_0|}} -\frac{E(T^{\alpha, \tau}_j)}{\sqrt{|S^c_0|}} \right] > n^{1/2}(1-p^{1-\beta-\gamma})^{1/2} n^{-r}(1+o(1))\right) \nonumber\\
&\leq & \exp\left\{ \frac{-2n^{1-2r}(1-p^{1-\beta-\gamma})(1+o(1))}{\sum_{j\in S_0^c}\left(\frac{1}{\sqrt{|S^c_0|}}\right)^2}\right\}\rightarrow 0 \mbox{ as } n\rightarrow \infty. \qed\nonumber
\end{eqnarray*}

\noindent\textbf{Proof of Theorem \ref{prop:Aymp_OPT}.}
Since the proof of (T\ref{prop:Aymp_OPT}.1) can also be used to show Theorem \ref{prop:postMR}, we only provide details for proving (T\ref{prop:Aymp_OPT}.1). For (T\ref{prop:Aymp_OPT}.2) and (T\ref{prop:Aymp_OPT}.3). Their proofs are straightforward and therefore are omitted here. Throughout the proofs of Theorems \ref{prop:Aymp_OPT} and \ref{prop:postMR}, we use the notation $\tilde{\boldsymbol{\Sigma}}^{-1}=\hat{\boldsymbol{\Omega}}^{\alpha, \tau}$ for the precision matrix estimated from the CIS procedure. Denote $\rho_n =M_0 C_{h,p}( n^{-1}\log p )^{(1-h)/2}$, then by (\ref{eqn:precisionBound}),
\begin{equation}\label{eqn:Denominator}
\hat{\boldsymbol{\delta}}' \tilde{\boldsymbol{\Sigma}}^{-1} \boldsymbol{\Sigma} \tilde{\boldsymbol{\Sigma}}^{-1}  \hat{\boldsymbol{\delta}}=\hat{\boldsymbol{\delta}}' \tilde{\boldsymbol{\Sigma}}^{-1}\hat{\boldsymbol{\delta}}(1+O_P(\rho_n))= \hat{\boldsymbol{\delta}}' \boldsymbol{\Sigma}^{-1}\hat{\boldsymbol{\delta}}(1+O_P(\rho_n)).
\end{equation}
For $\hat{\boldsymbol{\delta}}' \boldsymbol{\Sigma}^{-1}\hat{\boldsymbol{\delta}}$, notice that
\begin{equation*}
\hat{\boldsymbol{\delta}}' \boldsymbol{\Sigma}^{-1}\hat{\boldsymbol{\delta}}=\Delta_p^2 +2\boldsymbol{\delta}' \boldsymbol{\Sigma}^{-1}(\hat{\boldsymbol{\delta}}-\boldsymbol{\delta})+ (\hat{\boldsymbol{\delta}}-\boldsymbol{\delta})'\boldsymbol{\Sigma}^{-1}(\hat{\boldsymbol{\delta}}-\boldsymbol{\delta}),
\end{equation*}
where $\hat{\boldsymbol{\delta}}-\boldsymbol{\delta}=(\hat{\boldsymbol{\mu}}_1-\boldsymbol{\mu}_1)-(\hat{\boldsymbol{\mu}}_2-\boldsymbol{\mu}_2)$. Denote by $(\hat{\boldsymbol{\delta}}-\boldsymbol{\delta})'\boldsymbol{\Sigma}^{-1}(\hat{\boldsymbol{\delta}}-\boldsymbol{\delta})|_{S_0}=(\hat{\boldsymbol{\delta}}-\boldsymbol{\delta})_{S_0}'\boldsymbol{\Sigma}_{S_0}^{-1}(\hat{\boldsymbol{\delta}}-\boldsymbol{\delta})_{S_0}$ the sum of quadratic terms from feature in $S_0$, and denote by $(\hat{\boldsymbol{\delta}}-\boldsymbol{\delta})'\boldsymbol{\Sigma}^{-1}(\hat{\boldsymbol{\delta}}-\boldsymbol{\delta})|_{S^c_0}$ the sum of quadratic terms from feature in $S^c_0$. Then on $S_0$, $(\hat{\boldsymbol{\delta}}-\boldsymbol{\delta})_{S_0}\sim N(0, \frac{n}{n_1n_2}\boldsymbol{\Sigma}_{S_0})$. Assume the singular value decomposition of $\boldsymbol{\Sigma}_{S_0} =\mathbf{D} \boldsymbol{\Lambda}_{S_0} \mathbf{D}'$, where $\mathbf{D}$ is an orthogonal matrix and $\boldsymbol{\Lambda}_{S_0}=\mbox{diag}(\lambda_1, \cdots, \lambda_{|S_0|})$ are the eigenvalues of $\boldsymbol{\Sigma}_{S_0}$. Then
\[\boldsymbol{\epsilon}\equiv\sqrt{\frac{n_1n_2}{n}} \boldsymbol{\Lambda}_{S_0}^{-\frac{1}{2}} \mathbf{D}' (\hat{\boldsymbol{\delta}}-\boldsymbol{\delta})_{S_0} \sim N(0, \mathbf{I}_{|S_0|}).\]
And
\begin{equation}\label{eqn:Chi2}
(\hat{\boldsymbol{\delta}}-\boldsymbol{\delta})'\boldsymbol{\Sigma}^{-1}(\hat{\boldsymbol{\delta}}-\boldsymbol{\delta})|_{S_0}= \frac{n}{n_1n_2}\boldsymbol{\epsilon}' \boldsymbol{\epsilon}.
\end{equation}
By the weak law of large number,
\[\frac{n_1n_2}{p^{2-\beta-\gamma}n}(\hat{\boldsymbol{\delta}}-\boldsymbol{\delta})'\boldsymbol{\Sigma}^{-1}(\hat{\boldsymbol{\delta}}-\boldsymbol{\delta})|_{S_0} \stackrel{P}{\longrightarrow} 1 \ \ \ \mbox{ as }\ n\rightarrow \infty, \ p\rightarrow\infty.\]
The term $p^{2-\beta-\gamma}$ in the above equation corresponds to the marginally informative and MUJI features in ${S_0}$. On the other hand, $(\hat{\boldsymbol{\delta}}-\boldsymbol{\delta})'\boldsymbol{\Sigma}^{-1}(\hat{\boldsymbol{\delta}}-\boldsymbol{\delta})|_{S_0^c} \leq \kappa_2\sum_{j\in S_0^c} |\hat{\delta}_j-\delta_j|^2\leq \kappa_2\hat{b}_n |\hat{\delta}_j-\delta_j|^2$, where $\hat{b}_n=$ the number of $j$'s with $|\hat\delta_j|>\tau$. The last step inequality stems from that in $S_0^c$, $|\hat{\delta}_j-\delta_j|$ is nonzero only when $\hat{\delta}_j$ is not estimated as zero. By Lemma 2 in \citet{ShaoJ:2011}, $\hat{b}_n-b_n \stackrel{P}{\rightarrow} 0$. Notice that $\sqrt{n}(\hat{\delta}_j-\delta_j)\sim N(0,\sigma_j)$, so $|\hat{\boldsymbol{\delta}}_j-\boldsymbol{\delta}_j|^2 =O_P(n^{-1})$ for any entry $j$ of $\hat{\boldsymbol{\delta}}$ and as a result, $(\hat{\boldsymbol{\delta}}-\boldsymbol{\delta})'\boldsymbol{\Sigma}^{-1}(\hat{\boldsymbol{\delta}}-\boldsymbol{\delta})|_{S_0^c} \leq O_P(b_n n^{-1})$. Therefore,
\begin{eqnarray*}
\lefteqn{\frac{n_1n_2}{p^{2-\beta-\gamma}n}(\hat{\boldsymbol{\delta}}-\boldsymbol{\delta})'\boldsymbol{\Sigma}^{-1}(\hat{\boldsymbol{\delta}}-\boldsymbol{\delta})}\\
&=&\frac{n_1n_2}{p^{2-\beta-\gamma}n}\left\{(\hat{\boldsymbol{\delta}}-\boldsymbol{\delta})'\boldsymbol{\Sigma}^{-1}(\hat{\boldsymbol{\delta}}-\boldsymbol{\delta})|_{S_0}+(\hat{\boldsymbol{\delta}}-\boldsymbol{\delta})'\boldsymbol{\Sigma}^{-1}(\hat{\boldsymbol{\delta}}-\boldsymbol{\delta})|_{S_0^c}\right\}\\
&=& 1+ \frac{n_1n_2}{p^{2-\beta-\gamma}n}\Delta_p^2O_P(b_n/(\Delta_p^2n)).
\end{eqnarray*}
Similarly to the above arqument,
\[\boldsymbol{\delta}' \boldsymbol{\Sigma}^{-1}(\hat{\boldsymbol{\delta}}-\boldsymbol{\delta})\leq \Delta_p\sqrt{(\hat{\boldsymbol{\delta}}-\boldsymbol{\delta})'\boldsymbol{\Sigma}^{-1}(\hat{\boldsymbol{\delta}}-\boldsymbol{\delta})}=\Delta_p O_P(b_n^{1/2}n^{-1/2}).\]
Therefore
\begin{eqnarray}\label{eqn:nume_NEW}
&& \frac{n_1n_2}{p^{2-\beta-\gamma}n}\hat{\boldsymbol{\delta}}' \boldsymbol{\Sigma}^{-1}\hat{\boldsymbol{\delta}} \\
&=&1+ \frac{n_1n_2}{p^{2-\beta-\gamma}n}\Delta_p^2\left(1+O_P(b_n/(\Delta_p^2n))+O_P(b_n^{1/2}/(\Delta_pn^{1/2}))\right)\nonumber\\
&=& 1+ \frac{n_1n_2}{p^{2-\beta-\gamma}n}\Delta_p^2\left(1+O_P(b_n^{1/2}/(\Delta_pn^{1/2}))\right). \nonumber
\end{eqnarray}
Next we look at the numerator in the normal cumulative density function in (\ref{eqn:MCerror}). Similar to \ref{eqn:Denominator},
\[(\boldsymbol{\mu}_k- \hat{\boldsymbol{\mu}})'\tilde{\boldsymbol{\Sigma}}^{-1} \hat{\boldsymbol{\delta}}=(\boldsymbol{\mu}_k- \hat{\boldsymbol{\mu}})'\boldsymbol{\Sigma}^{-1} \hat{\boldsymbol{\delta}} (1+O_P(\rho_n)), \ \ \ k=1,2.\]
And straight forward calculation gives that
\begin{eqnarray}
(\boldsymbol{\mu}_1- \hat{\boldsymbol{\mu}})'\boldsymbol{\Sigma}^{-1} \hat{\boldsymbol{\delta}}&=&\frac{1}{2}\boldsymbol{\delta}'\boldsymbol{\Sigma}^{-1}\boldsymbol{\delta} - \boldsymbol{\delta}' \boldsymbol{\Sigma}^{-1}(\hat{\boldsymbol{\mu}}_2 - \boldsymbol{\mu}_2) -\frac{1}{2}(\hat{\boldsymbol{\mu}}_1 - \boldsymbol{\mu}_1)'\boldsymbol{\Sigma}^{-1}(\hat{\boldsymbol{\mu}}_1 - \boldsymbol{\mu}_1) +\nonumber\\
&&\frac{1}{2}(\hat{\boldsymbol{\mu}}_2 - \boldsymbol{\mu}_2)'\boldsymbol{\Sigma}^{-1}(\hat{\boldsymbol{\mu}}_2 - \boldsymbol{\mu}_2) \nonumber\\
&=& \frac{1}{2} \Delta_P^2 -I_1- \frac{1}{2}I_2 -\frac{1}{2}I_3, \label{eqn:I1I2I3}
\end{eqnarray}
where $I_1\equiv \boldsymbol{\delta}' \boldsymbol{\Sigma}^{-1}(\hat{\boldsymbol{\mu}}_2 - \boldsymbol{\mu}_2)$, $I_2 \equiv (\hat{\boldsymbol{\mu}}_1 - \boldsymbol{\mu}_1)'\boldsymbol{\Sigma}^{-1}(\hat{\boldsymbol{\mu}}_1 - \boldsymbol{\mu}_1)$ and $I_3 \equiv (\hat{\boldsymbol{\mu}}_2 - \boldsymbol{\mu}_2)'\boldsymbol{\Sigma}^{-1}(\hat{\boldsymbol{\mu}}_2 - \boldsymbol{\mu}_2)$. Notice that $\boldsymbol{\delta}'\boldsymbol{\Sigma}^{-1}(\hat{\boldsymbol{\mu}}_2 - \boldsymbol{\mu}_2)\sim N(0, \frac{1}{n_2}\boldsymbol{\delta}'\boldsymbol{\Sigma}^{-1}\boldsymbol{\delta})$. As a result,
\[I_1=\frac{1}{2}\Delta_p^2 o_P(1).\]
Since $\hat{\boldsymbol{\mu}}_1-\boldsymbol{\mu}_1\sim N(0, \frac{1}{n_1}\boldsymbol{\Sigma})$, define $\tilde{\epsilon}_1=\sqrt{n_1}\Lambda^{-1/2}\mathbf{D}'(\hat{\boldsymbol{\mu}}_1-\boldsymbol{\mu}_1)$, then $\tilde{\epsilon}_1\sim N(0,\mathbf{I})$ and $I_2=\frac{1}{n_1}\tilde{\epsilon}'_1\tilde{\epsilon}_1$. Therefore, by the weak law of large numbers,
\[\frac{n_1}{p} I_2 \stackrel{P}{\longrightarrow} 1.\]
On the other hand, $\mbox{Var}(I_2)\leq \mbox{Var}((\hat{\boldsymbol{\delta}}_1-\boldsymbol{\delta}_1)'\boldsymbol{\Sigma}^{-1}(\hat{\boldsymbol{\delta}}_1-\boldsymbol{\delta}_1))$. And from (\ref{eqn:Chi2}), $(n_1n_2)/(pn)[(\hat{\boldsymbol{\delta}}-\boldsymbol{\delta})'\boldsymbol{\Sigma}^{-1}(\hat{\boldsymbol{\delta}}-\boldsymbol{\delta})]\sim \chi_1^2$. Therefore, $\mbox{Var}(I_2)=o_p\left(pn/(n_1 n_2)\right)$ and $I_2=\frac{p}{n_1} +o_P\left(\sqrt{\frac{pn}{n_1 n_2}}\right)$. Moreover,
\begin{eqnarray*}
&& \boldsymbol{\delta}'\boldsymbol{\Sigma}^{-1}(\hat{\boldsymbol{\mu}}_2 - \boldsymbol{\mu}_2)=\frac{1}{2}\Delta_p^2 o_P(1), \ \ \ \ \frac{1}{2}\boldsymbol{\Sigma}^{-1}(\hat{\boldsymbol{\mu}}_1 - \boldsymbol{\mu}_1)^2=\frac{p}{n_1} +o_P\left(\sqrt{\frac{pn}{n_1 n_2}}\right) \\
&& \mbox{ and } \frac{1}{2}\boldsymbol{\Sigma}^{-1}(\hat{\boldsymbol{\mu}}_2 - \boldsymbol{\mu}_2)^2=\frac{p}{n_2} +o_P\left(\sqrt{\frac{pn}{n_1 n_2}}\right).
\end{eqnarray*}
Plugging these into (\ref{eqn:I1I2I3}), we have
\begin{equation}\label{eqn:numo_NEW}
(\boldsymbol{\mu}_k- \hat{\boldsymbol{\mu}})'\boldsymbol{\Sigma}^{-1} \hat{\boldsymbol{\delta}}=\frac{1}{2}\Delta_p^2(1+o_P(1))+\frac{p(n_1-n_2)}{2n_1n_2}+o_P\left(\sqrt{\frac{pn}{n_1 n_2}}\right).
\end{equation}
Combining (\ref{eqn:nume_NEW}) and (\ref{eqn:numo_NEW}), we get
\begin{eqnarray}\label{eqn:quot}
&&\frac{(\boldsymbol{\mu}_k- \hat{\boldsymbol{\mu}})'\hat{\boldsymbol{\Sigma}}^{-1} \hat{\boldsymbol{\delta}}}{\sqrt{\hat{\boldsymbol{\delta}}' \tilde{\boldsymbol{\Sigma}}^{-1} \boldsymbol{\Sigma} \tilde{\boldsymbol{\Sigma}}^{-1}  \hat{\boldsymbol{\delta}}}} \\
&=&\frac{\left\{\sqrt{\frac{n_1n_2}{pn}}\Delta_p^2(1+o_P(1))+ \sqrt{\frac{p}{nn_1n_2}}(n_1-n_2)+o_p(1)\right\}(1+O_P(\rho_n))}{2 \left\{\left[1+\frac{n_1n_2}{p^{2-\beta-\gamma}n}\Delta_p^2\left(1+O_P\left(\frac{b_n^{1/2}}{\Delta_pn^{1/2}}\right)\right)\right](1+O_P(\rho_n))\right\}^{1/2}}\nonumber\\
&=&\frac{\left\{\Delta_p\left[\sqrt{p^{1-\beta-\gamma}}(1+o_P(1))+ d_n\right]+o_p(1)\right\}(1+O_P(\rho_n))^{1/2}}{2 \left\{1+O_P\left(\frac{b_n^{1/2}}{\Delta_pn^{1/2}}\right)\right\}^{1/2}}, \nonumber
\end{eqnarray}
where $d_n=p^{\frac{3-\beta-\gamma}{2}}(n_1-n_2)/(n_1n_2\Delta_p^2)=O_P(a_n)$. Under the assumptions (A6)-(A10), it is easy to see that
\[\frac{(\boldsymbol{\mu}_k- \hat{\boldsymbol{\mu}})'\tilde{\boldsymbol{\Sigma}}^{-1} \hat{\boldsymbol{\delta}}}{\sqrt{\hat{\boldsymbol{\delta}}' \tilde{\boldsymbol{\Sigma}}^{-1} \boldsymbol{\Sigma} \tilde{\boldsymbol{\Sigma}}^{-1}  \hat{\boldsymbol{\delta}}}}=\frac{\Delta_p}{2}[\sqrt{p^{1-\beta-\gamma}}+O_P(a_n)].  \qed\]

\noindent\textbf{Proof of Theorem \ref{prop:postMR}.} Notice that (\ref{eqn:quot}) holds for any $p$.
When post-screening classification is based only on the selected feature set $S^{CIS}(\alpha, \tau; \nu_n)$, by replacing $p$ with $|S^{CIS}(\alpha, \tau; \nu_n)|$ and $p^{2-\beta-\gamma}$ with $O_P(|S^{CIS}(\alpha, \tau; \nu_n)|)$ in the first equality in (\ref{eqn:quot}), we have
\[\frac{ (\boldsymbol{\mu}^{\cis}_k- \hat{\boldsymbol{\mu}}^{\cis})'\hat{\boldsymbol{\Omega}}^{\cis} \hat{\boldsymbol{\delta}}^{\cis}}{\sqrt{\hat{\boldsymbol{\delta}}^{\cis '} \hat{\boldsymbol{\Omega}}^{\cis} \boldsymbol{\Sigma}^{\cis}\hat{\boldsymbol{\Omega}}^{\cis}  \hat{\boldsymbol{\delta}}^{\cis}}}=\frac{\Delta_p}{2}[1+O_P(a_n)]. \qed\]

\noindent\textbf{Proof of Theorem \ref{prop:prop6}.}
We need to show that there exist $(\tau, \alpha, \nu_n)$ satisfying assumption (A5), such that the selected feature set with those thresholding parameters satisfies conditions (C1) and (C2).

Notice that (i) $j\in S_0$ if and only if $\mu_{1j}-\mu_{2j}\neq 0$ or $\mu_{1j}-\mu_{2j}= 0$ but $\Omega_{jj'} \neq 0$ for some $j'\neq j$ such that $\mu_{1j'}-\mu_{2j'}\neq 0$ and (ii) $j\in S_0^c$ if and only if $\mu_{1j}-\mu_{2j}=0$ and $\Omega_{jj'}=0$ for all $j'\neq j$ such that $\mu_{1j'}-\mu_{2j'}\neq 0$. Based on these two facts, $S_{01}$ and $S_{02}$ can be re-expressed as $S_{01}=\{j\in S_0, \mu_{1j}-\mu_{2j}\neq 0\}$ the subset of $S_0$ with non-zero mean difference between the two classes, $S_{02}=\{j\in S_0, \mu_{1j}-\mu_{2j}= 0, \Omega_{jj'} \neq 0 \mbox{ for some } j'\neq j, \mbox{ s.t. } \mu_{1j'}-\mu_{2j'}\neq 0\}$ the subset of $S_0$ with zero mean difference but is correlated with some non-zero mean difference features. Then $S_{01}$ and $S_{02}$ are mutually exclusive.

Denote by ETP the expected true positives, i.e., the expected number of indices in the set $\{j \in S_0: \tilde{T}_j\neq 0, j=1,\cdots,p \}$ and EFP the expected false positives, i.e., the expected number of indices in the set $\{j \in S_0^c: \tilde{T}_j\neq 0, j=1,\cdots,p \}$ with $\tilde{T}_j\equiv\sum_{j'=1}^p \hat{\boldsymbol{\Omega}}_{jj'}(\hat{\mu}_{1j'}- \hat{\mu}_{2j'})\equiv \sum_{j'=1}^p \hat{\boldsymbol{\Omega}}_{jj'}\hat\delta_j$.

For simplicity, assume $\boldsymbol{\mu}_1=\mathbf{0}$ and an equal variance, i.e., the diagonal terms of $\boldsymbol{\Sigma}$ are all equal to $\sigma^2$. According to the CIS procedure, the set of true positives is
\[\mbox{TP set} = \left\{j: |\hat\delta_j|> \tau, j \in S_{01}\right\} \cup \left\{j: |\tilde{T}_j| > \nu_n, j \in S_{02}\cap S_{01}^c\right\}.\]
Therefore,
\begin{eqnarray*}
&& P(\mbox{TP set}) \\
&=& P\left(|\hat\delta_j|> \tau\, |\, j \in S_{01}\right)P(S_{01}) + P\left(|\tilde{T}_j| > \nu_n\, |\, j \in S_{02}\cap S_{01}^c\right) P(S_{02}\cap S_{01}^c)\\
&=& \epsilon P\left(|\hat\delta_j|> \tau\, |\, S_{01}\right) + (1-\epsilon)\vartheta P\left(|\tilde{T}_j| > \nu_n\, |\, S_{02}\cap S_{01}^c\right).
\end{eqnarray*}
Notice that
\begin{equation}\label{eqn:PofTau1}
P\left(|\hat\delta_j|> \tau\, |\, S_{01}\right)=P\left(Z>\frac{\tau-\mu_{2j}}{\sigma} \right) + P\left(Z<\frac{-\tau-\mu_{2j}}{\sigma} \right),
\end{equation}
and $P\left(|\tilde{T}_j| > \nu_n\, |\, S_{02}\cap S_{01}^c\right)= P\left(|(\hat{\boldsymbol{\Omega}}\hat{\boldsymbol{\delta}})_j| > \nu_n \, |\,  S_{02}\cap S_{01}^c\right)$. From Lemma \ref{lem:lemE},
\begin{equation*}
P\left(|(\hat{\boldsymbol{\Omega}}\hat{\boldsymbol{\delta}})_j| > \nu_n \, |\,  j\in S_{02}\cap S_{01}^c\right)\geq P\left(\|\hat{\boldsymbol{\delta}}_{\cdot \hat{\mathcal{E}}_{j\alpha}}\|_1 > \kappa_1^{-1} \nu_n\right)\geq P\left(|\hat{\delta}_{j'}| > \kappa_1^{-1} \nu_n \right)
\end{equation*}
with some $j'\in S_{01}$ and $\Omega_{jj'}\neq 0$. Therefore, $\nu_n$ can be chosen such that $\kappa_1^{-1} \nu_n \sim \tau$, and
\begin{equation}\label{eqn:TPP}
 P(\mbox{TP set}) \geq \left(\epsilon+\vartheta\right)P\left(|\hat\delta_j|> \tau\, |\, S_{01}\right)\left(1-o_P(1)\right).
\end{equation}
Similarly for the false positive set
\[\mbox{FP set} = \left\{j: |\hat\delta_j|> \tau, j \in S_{01}^c\right\} \cup \left\{j: |\tilde{T}_j| > \nu_n, j \in S_{02}^c\cap S_{01}^c\right\}.\]
\begin{equation*}
P(\mbox{FP set})= (1-\epsilon) P\left(|\hat\delta_j|> \tau \,|\, S_{01}^c\right) + (1-\epsilon)(1-\vartheta) P\left(|\tilde{T}_j| > \nu_n \,|\, S_{02}^c\cap S_{01}^c\right)
\end{equation*}
with
\begin{equation}\label{eqn:PofTau2}
P\left(|\hat\delta_j| > \tau \,|\, S_{01}^c\right)=P\left(Z > \frac{\tau}{\sigma} \right)+P\left(Z < \frac{-\tau}{\sigma} \right),
\end{equation}
and follows a similar argument as above,
\[P\left(|\tilde{T}_j| > \nu_n \,|\, S_{02}^c\cap S_{01}^c \right)\leq P\left(|\hat\delta_{j'}| > \tau \,|\, S_{01}^c \right),\]
for some $1\leq j'\leq p$. Therefore, for the false positives, we have
\begin{equation}\label{eqn:FPP}
(1-\epsilon) P\left(|\hat\delta_j| > \tau \,|\, S_{01}^c\right) \leq P(\mbox{FP set}) \leq 2(1-\epsilon) P\left(|\hat\delta_j| > \tau \,|\, S_{01}^c\right).
\end{equation}
From (\ref{eqn:TPP}) and (\ref{eqn:FPP}), we have
\begin{equation}\label{eqn:ETP}
\mbox{ETP} \geq p(\epsilon+\vartheta) P\left(|\hat\delta_j|> \tau\, |\, S_{01}\right)\left(1-o_P(1)\right), \ \ \ \mbox{and}
\end{equation}
\begin{equation}\label{eqn:FPP1}
p(1-\epsilon) P\left(|\hat\delta_j| > \tau \,|\, S_{01}^c\right)\leq \mbox{EFP} \leq p(1-\epsilon) 2 P\left(|\hat\delta_j| > \tau \,|\, S_{01}^c\right).
\end{equation}
Under the assumption $\beta>\gamma$, $\epsilon+\vartheta=p^{-\beta} + p^{-\gamma}=p^{-\gamma}(1+p^{-(\beta-\gamma)})\sim p^{-\gamma}=p^{-\pi\beta}$, where $\pi = \gamma/\beta \in(0,1)$. Therefore $\mbox{ETP}\geq p^{1-\pi\beta} P\left(|\hat\delta_j|> \tau\, |\, S_{01}\right)$. And $\mbox{EFP}\leq 2p(1-p^{-\beta}) P\left(|\hat\delta_j| > \tau \,|\, S_{01}^c\right)$.

First consider the decision rule $\mathbbm{1}(\delta_j>\tau)$. For decision rules $\mathbbm{1}(\delta_j<-\tau)$ and $\mathbbm{1}(|\delta_j|>\tau)$, they follow a similar argument with $\mathbbm{1}(\delta_j>\tau)$. We only discuss the case when $0<\sigma<1$. Conclusions follow with similar arguments for the cases when $\sigma=1$ and $\sigma>1$ and the detailed proof will be provided in the supplemental material. When $r>(1-\sigma\sqrt{1-\pi\beta})^2$, for the following threshold of order $O(\sqrt{r\log p})$ with some $\epsilon_0\in(0,1)$:
\[\tau^*=\left\{\frac{\sqrt{r}-\sigma\sqrt{r-\pi\beta(1-\sigma^2)}+\sigma\epsilon_0}{1-\sigma^2}\right\}\sqrt{2\log p}.\]
$\tau^*$ is well defined as it is easy to show that $(1-\sigma\sqrt{1-\pi\beta})^2 \geq (1-\sigma^2)\pi\beta$ when $0<\sigma<1$ and $0<\pi\beta<1$. Straight calculation shows that for any $1\leq j \leq p$,
\begin{eqnarray*}
\frac{\tau^*-\mu_{2j}}{\sigma}&=& \left.\left\{\frac{\sqrt{r}-\sigma\sqrt{r-\pi\beta(1-\sigma^2)}+\sigma\epsilon_0}{1-\sigma^2}-\sqrt{r}\right\}\sqrt{2\log p}\right/\sigma\\
&=& \left\{\frac{\sigma\sqrt{r}-\sqrt{r-\pi\beta(1-\sigma^2)}+\epsilon_0}{1-\sigma^2}\right\}\sqrt{2\log p}.
\end{eqnarray*}
And
\begin{equation*}
\frac{\tau^*}{\sigma}\geq \tau^*= \left\{\frac{\sqrt{r}-\sigma\sqrt{r-\pi\beta(1-\sigma^2)}+\sigma\epsilon_0}{1-\sigma^2}\right\}\sqrt{2\log p}.
\end{equation*}
Then from the Gaussian tail approximation,
\begin{eqnarray*}
&&\mbox{ETP}\geq \frac{p^{1-\pi\beta} (1-\sigma^2)}{2\sqrt{\pi\log p} \{\sigma\sqrt{r}-\sqrt{r-\pi\beta(1-\sigma^2)} + \epsilon_0\}}
p^{-\left\{\frac{\sigma\sqrt{r}-\sqrt{r-\pi\beta(1-\sigma^2)}+\epsilon_0}{1-\sigma^2}\right\}^2},\nonumber\\
&&\mbox{EFP}\leq  \frac{p(1-p^{-\beta})(1-\sigma^2)}{\sqrt{\pi\log p} \{\sqrt{r}-\sigma\sqrt{r-\pi\beta(1-\sigma^2)} + \sigma\epsilon_0\}}
p^{-\left\{\frac{\sqrt{r}-\sigma\sqrt{r-\pi\beta(1-\sigma^2)}+\sigma\epsilon_0}{1-\sigma^2}\right\}^2}.
\end{eqnarray*}
Then
\begin{eqnarray*}
\frac{\mbox{ETP}}{\mbox{EFP}}&\geq& \frac{\sqrt{r}+\sigma\epsilon_0-\sigma\sqrt{r-\pi\beta(1-\sigma^2)}}{2(\sigma\sqrt{r}+\epsilon_0-\sqrt{r-\pi\beta(1-\sigma^2)})}\times\\
&& p^{-\pi\beta-\left\{\frac{\sigma\sqrt{r}-\sqrt{r-\pi\beta(1-\sigma^2)}+\epsilon_0}{1-\sigma^2}\right\}^2+\left\{\frac{\sqrt{r}-\sigma\sqrt{r-\pi\beta(1-\sigma^2)}+\sigma\epsilon_0}{1-\sigma^2}\right\}^2}\\
&=& \frac{\sqrt{r}+\sigma\epsilon_0-\sigma\sqrt{r-\pi\beta(1-\sigma^2)}}{2(\sigma\sqrt{r}+\epsilon_0-\sqrt{r-\pi\beta(1-\sigma^2)})}p^{\frac{\epsilon_0\{2\sqrt{r-\pi\beta(1-\sigma^2)}-\epsilon_0\}}{1-\sigma^2}}\rightarrow \infty.
\end{eqnarray*}
when chosing $\epsilon_0< 2\sqrt{r-\pi\beta(1-\sigma^2)}$.

We next show that there exists an $\epsilon_0$ such that with a probability tending to $1$, the discovery set is nonempty. Let $\xi_p=P(X_j>\tau^*)$, then $p\xi_p=\mbox{ETP} + \mbox{EFP}$. It can be easily shown that
\[ \pi\beta + \left\{\frac{\sigma\sqrt{r}-\sqrt{r-\pi\beta(1-\sigma^2)}}{1-\sigma^2}\right\}^2 = \left\{\frac{\sqrt{r}-\sigma\sqrt{r-\pi\beta(1-\sigma^2)}}{1-\sigma^2}\right\}^2,\]
which gives that
\[\xi_p=Cp^{-\pi\beta-\left\{\frac{\sigma\sqrt{r}-\sqrt{r-\pi\beta(1-\sigma^2)}+\epsilon_0}{1-\sigma^2}\right\}^2},\]
for some positive constant $C <2$. Similar to Lemma 7 in \citep{CaiSun:2014}, it is easy to show that $\pi\beta+\left\{\frac{\sigma\sqrt{r}-\sqrt{r-\pi\beta(1-\sigma^2)}}{1-\sigma^2}\right\}^2 <1$ when $r> d^{\cis}(\beta)$. Therefore there exists an $\epsilon_0<2\sqrt{r-\pi\beta(1-\sigma^2)}$ such that $\pi\beta+\left\{\frac{\sigma\sqrt{r}+\epsilon_0-\sqrt{r-\pi\beta(1-\sigma^2)}}{1-\sigma^2}\right\}^2 <1$. And further there exists a $\kappa >0$ such that $\pi\beta+\left\{\frac{\sigma\sqrt{r}-\sqrt{r-\pi\beta(1-\sigma^2)}+\epsilon_0}{1-\sigma^2}\right\}^2<1-\kappa$.
Then the probability of having a non-empty discovery set is
\[P(|S_\delta|\geq 1)=1-(1-\xi_p)^p > 1-(1-p^{-(1-\kappa)})^p > 1-e^{-p^k}(1+o(1))\rightarrow 1.\]
\qed

\section*{Acknowledgements}
And this is an acknowledgements section with a heading that was produced by the
$\backslash$section* command. Thank you all for helping me writing this
\LaTeX\ sample file. See \ref{suppA} for the supplementary material example.

\begin{supplement}
\sname{Supplement A}\label{suppA}
\stitle{Title of the Supplement A}
\slink[url]{http://www.e-publications.org/ims/support/dowload/imsart-ims.zip}
\sdescription{Dum esset rex in
accubitu suo, nardus mea dedit odorem suavitatis. Quoniam confortavit
seras portarum tuarum, benedixit filiis tuis in te. Qui posuit fines tuos}
\end{supplement}

\bibliographystyle{imsart-nameyear}
\bibliography{mybib}
\end{document}